\pdfoutput=1

\documentclass[11pt]{article}

\usepackage[]{EMNLP2022}

\usepackage{times}
\usepackage{latexsym}
\usepackage{amsfonts,amssymb}
\usepackage{subfigure}
\usepackage{amsthm}
\usepackage[T1]{fontenc}
\usepackage[utf8]{inputenc}
\usepackage{graphicx}

\usepackage{amsmath}
\usepackage{bbm}
\usepackage{enumitem}
\usepackage{booktabs}
\usepackage{multirow}
\usepackage{microtype}
\usepackage{CJKutf8}
\usepackage{inconsolata}

\newcommand{\caomei}{\includegraphics[width=0.02\textwidth]{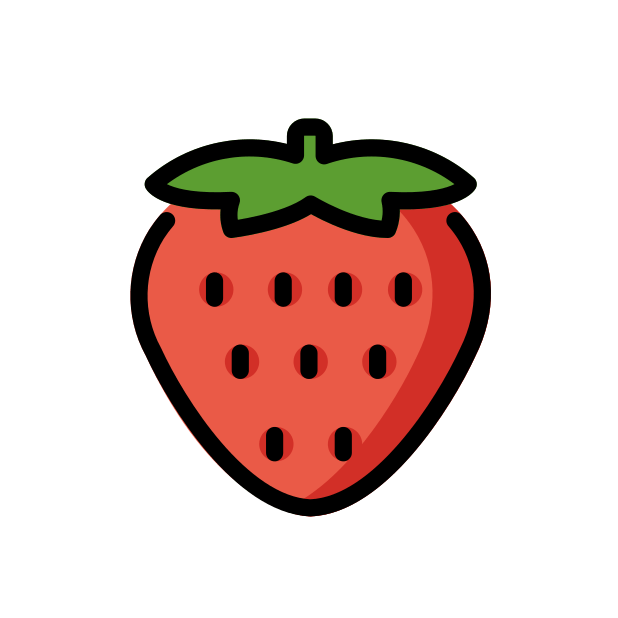}}
\newcommand{\mangguo}{\includegraphics[width=0.02\textwidth]{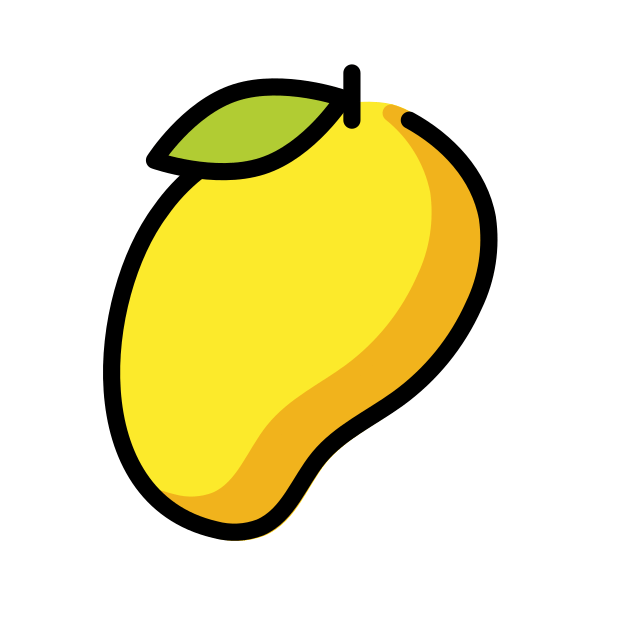}}
\newcommand{\putao}{\includegraphics[width=0.02\textwidth]{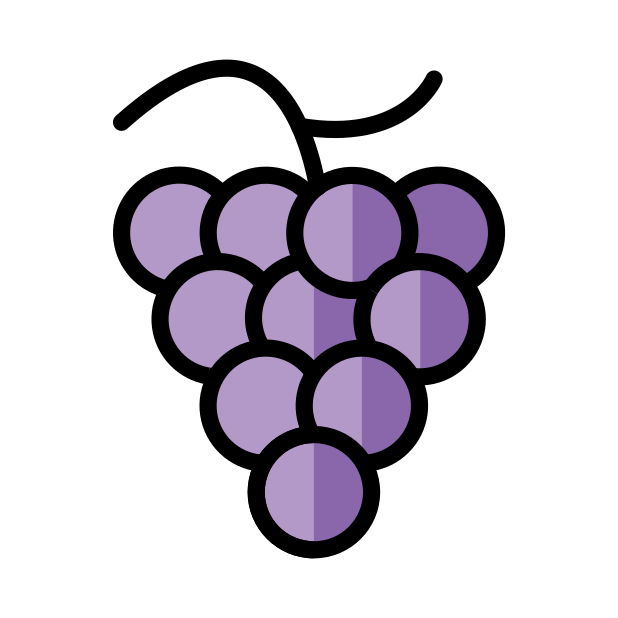}}
\newcommand{\pingguo}{\includegraphics[width=0.02\textwidth]{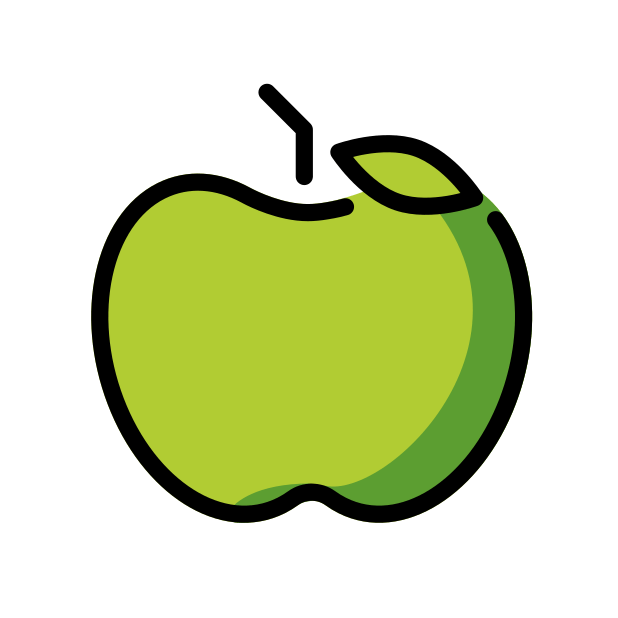}}
\newcommand{\xinfeng}{\includegraphics[width=0.02\textwidth]{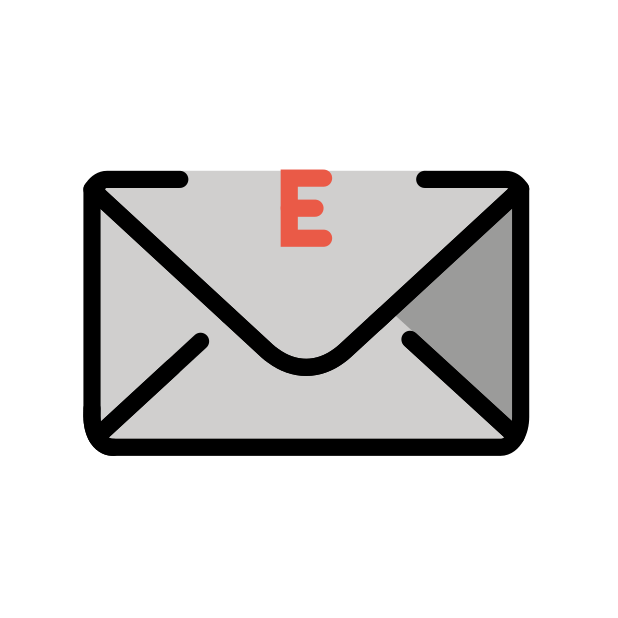}}

\title{\textsc{XPrompt}: Exploring the Extreme of Prompt Tuning}

\author{
  Fang Ma\textsuperscript{\caomei}, Chen Zhang\textsuperscript{\caomei}, Lei Ren\textsuperscript{\mangguo}, Jingang Wang\textsuperscript{\mangguo\xinfeng}, Qifan Wang\textsuperscript{\putao}, \\
  \textbf{Wei Wu\textsuperscript{\mangguo}, Xiaojun Quan\textsuperscript{\pingguo}, Dawei Song\textsuperscript{\caomei\xinfeng}\Thanks{\textsuperscript{\xinfeng}Dawei Song and Jingang Wang are the corresponding authors.}} \\
  \textsuperscript{\caomei}Beijing Institute of Technology \quad
  \texttt{\{mfang,czhang,dwsong\}@bit.edu.cn} \\
  \textsuperscript{\mangguo}Meituan NLP \quad
  \texttt{\{wangjingang02,wuwei30\}@meituan.com}, \texttt{renlei\_work@163.com} \\
  \textsuperscript{\putao}Meta AI \quad \texttt{wqfcr@fb.com} \\
  \textsuperscript{\pingguo}Sun Yat-Sen University \quad
  \texttt{quanxj3@mail.sysu.edu.cn} \\
}
\makeatletter
\def\thanks#1{\protected@xdef\@thanks{\@thanks
    \protect\footnotetext{#1}}}
\makeatother

\begin{document}
\maketitle

\begin{abstract}

Prompt tuning learns soft prompts to condition frozen Pre-trained Language Models (PLMs) for performing downstream tasks in a parameter-efficient manner. While prompt tuning has gradually reached the performance level of fine-tuning as the model scale increases, there is still a large performance gap between prompt tuning and fine-tuning for models of moderate and small scales (typically less than $11$B parameters). In this paper, we empirically show that the trained prompt tokens can have a negative impact on a downstream task and thus degrade its performance. To bridge the gap, we propose a novel \textsc{Prompt} tuning model with an eXtremely small scale (\textsc{XPrompt}) under the regime of lottery tickets hypothesis. Specifically, \textsc{XPrompt} eliminates the negative prompt tokens at different granularity levels through a hierarchical structured pruning, yielding a more parameter-efficient prompt yet with a competitive performance. Comprehensive experiments are carried out on SuperGLUE tasks, and the extensive results indicate that \textsc{XPrompt} is able to close the performance gap at smaller model scales.


\end{abstract}

\section{Introduction}

Pre-trained Language Models (PLMs) have been widely applied and achieved a remarkable success in various NLP tasks \citep{devlin2019bert, raffel2020exploring, zhou2020towards} under the \textit{pre-train-then-fine-tune} paradigm \cite{liu2019roberta}. Despite of its compelling performance, fine-tuning is parameter-inefficient for large scale PLMs due to the fact that the memory footprint is proportional to the number of trainable parameters whose gradients and optimizer states need to be stored \citep{guo2020parameter}.

\begin{figure}[t]
  \centering
  \resizebox{0.4\textwidth}{!}{
  \includegraphics[width=\linewidth]{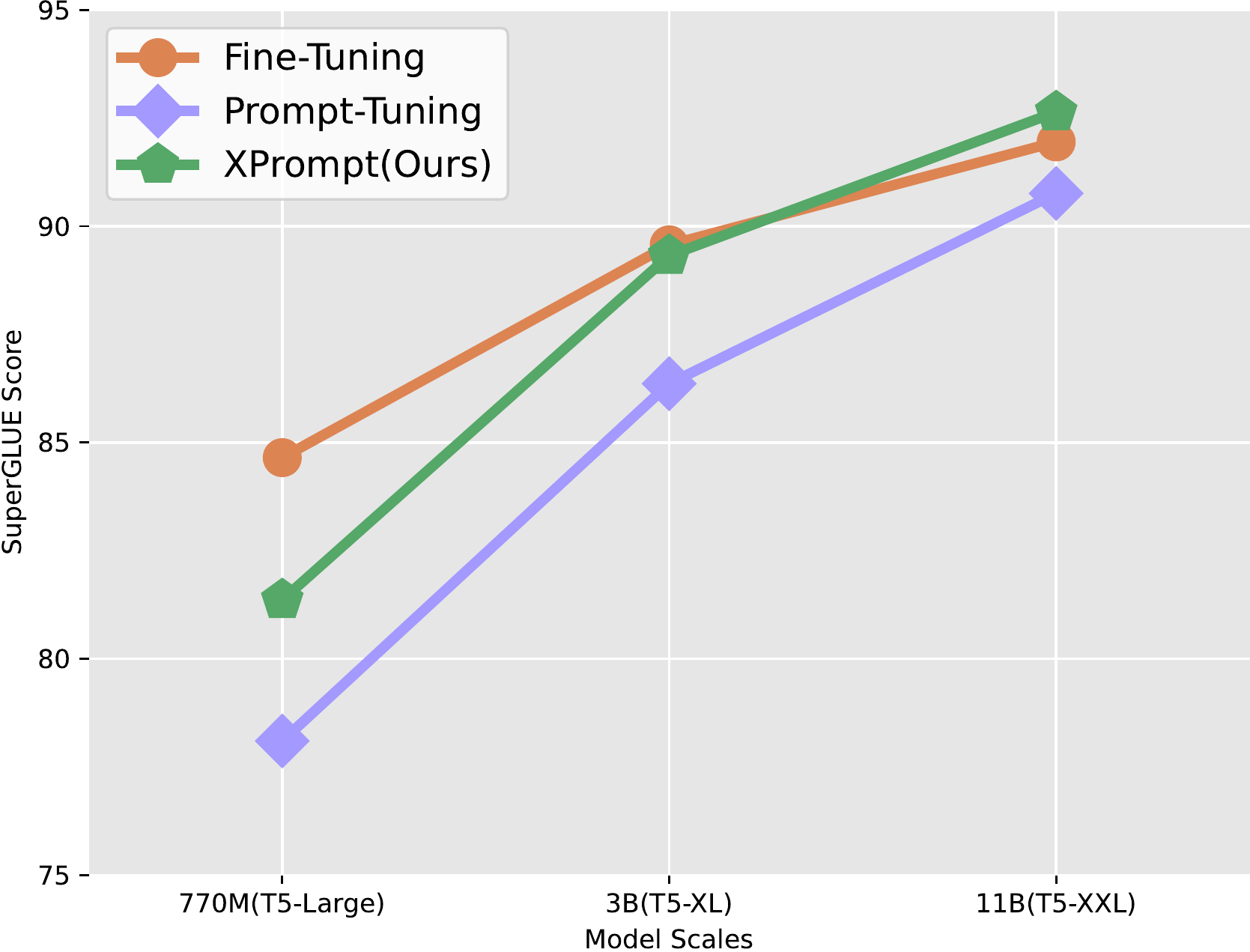}}
  \caption{\textsc{XPrompt} outperforms the vanilla Prompt-Tuning \citep{lester2021power} and can significantly improve over Prompt-Tuning across tasks and model scales. It is worth noting that there is a small performance gap between prompt tuning and fine-tuning on T$5$-XXL (11B) due to different hyperparameter settings and initialization. Similar observations have been found in Figure$3$-a and Figure$3$-b of \citet{lester2021power}.}
  \label{SuperGLUE_Line}
  \vspace{-5mm}
\end{figure}

Recently, Prompt-Tuning \citep{lester2021power,liu2021p} has been proposed to address this issue by prepending a \textit{soft prompt} to the input and only updating the parameters of prompt tokens during tuning. Prompt-Tuning provides a parameter-efficient alternative to fine-tuning, since the scale of the soft prompt is tens of thousand smaller. It is also conceptually simpler and more flexible than other parameter-efficient tuning methods such as Adapters that require intrusive modifications to transformer layers \citep{houlsby2019parameter, guo2020parameter}. Using fewer tunable parameters, prompt tuning achieves competitive performance to fine-tuning with the increase of the model scale. However, there is still a large performance gap between prompt tuning and fine-tuning for models of smaller scales (as shown in Figure~\ref{SuperGLUE_Line}).

\begin{figure}[ht]
  \centering
  \resizebox{0.4\textwidth}{!}{
  \includegraphics[width=\linewidth]{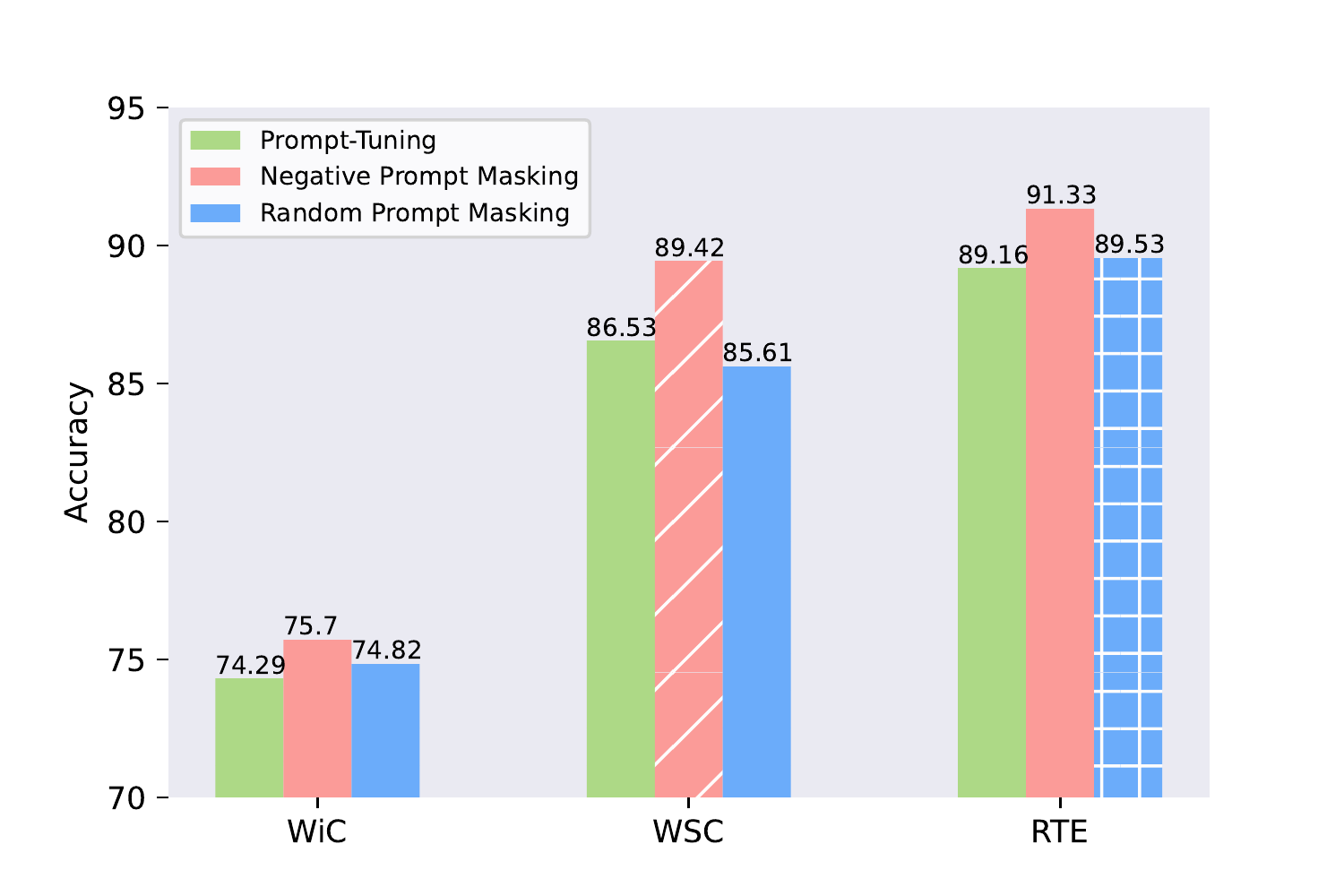}}
  \caption{The performance comparison of Prompt-Tuning, Negative Prompt Masking and Random Prompt Masking with T$5$-XL($3$B) on three SuperGLUE tasks. Prompt-Turning uses all prompt tokens. Negative Prompt Masking masks selected (negative) prompt tokens with low importance scores. Random Prompt Masking randomly masks the same number of tokens as in Negative Prompt Masking.}
  \label{intro_positive_negative_prompt}
  \vspace{-5mm}
\end{figure}

This paper aims to fill the gap, from the perspective of the lottery tickets hypothesis (LTH) \citep{frankle2018lottery}. We are motivated by an observation that, on a specific task, not all prompt tokens contribute equally to the task performance, while certain prompt tokens may even bring a negative impact. Figure~\ref{intro_positive_negative_prompt} provides a preliminary result of this observation.  These \textit{negative prompt tokens} can be circumvented  under the regime of LTH. Essentially, LTH states that an over-parameterized network contains a sub-network that, when initialized and trained in isolation, can match or exceed the test accuracy of the original network after training for at most the same number of iterations. The sub-network is called lottery ticket, and the collection of the tickets is referred to as winning tickets in PLMs \citep{liang2021super}. In the problem of prompt-tuning, the winning tickets are the collection of positive prompt tokens that can achieve the same performance as using the entire collection of prompts, while the losing tickets are the collection of negative prompt tokens.

Therefore, the key is to identify the winning tickets and eliminate the losing ones, in the collection of trained prompt tokens. 
In particular, we propose to eliminate the losing tickets through a hierarchical structured pruning, which first removes negative tokens at the token-level and then prunes the remaining ones  at a finer graularity level, i.e., the piece-level, for a better trade-off between effectiveness and efficiency. In line with LTH, weight rewinding \citep{renda2019comparing} is adopted to re-train the identified positive soft prompts. With the elimination of negative prompt tokens, a more parameter-efficient \textsc{Prompt} of an eXtremely small scale (\textsc{XPrompt}) is obtained.

To verify the effectiveness of \textsc{XPrompt}, we conduct an extensive set of experiments on SuperGLUE \citep{wang2019superglue} in both high-resource and low-resource scenarios. As shown in Figure~\ref{SuperGLUE_Line} and Table~\ref{mainSuperGLUEresults}, the results demonstrate that \textsc{XPrompt} significantly improves the prompt-tuning methods across tasks and model scales. For models of moderate scales, \textsc{XPrompt} closes the gap and achieves a performance comparable to fine-tuning. For models of large scales, \textsc{XPrompt} also leads to large performance gains over Prompt-Tuning, and even exceeds fine-tuning for most tasks.

\section{Related Work}

\subsection{Pre-trained Language Models}
Pre-trained Language Models (PLMs) have achieved remarkable success in various NLP tasks \citep{zhou2020towards, raffel2020exploring, brown2020language}. BERT \citep{devlin2019bert} and RoBERTa \citep{liu2019roberta} are two pioneers that learn contextual representations with masked language model (MLM) and next sentence prediction pre-training tasks. Recently, a series of large scale PLMs have emerged with different pre-training designs, such as GPT-$2$ \citep{radford2019language}, GPT-$3$ \citep{brown2020language}, ELECTRA \citep{clark2019electra}, XLNet \citep{yang2019xlnet}, BART \citep{lewis2020bart} and T$5$ \citep{raffel2020exploring}. However, with the exploding number of parameters, fine-tuning models become parameter-inefficient and computationally expensive due to the maintenance of all parameters in the PLMs. Moreover, one has to fine-tune different models for different tasks and store them separately, which is resource-intensive.

\begin{figure*}[ht]
  \centering
  \includegraphics[width=\linewidth]{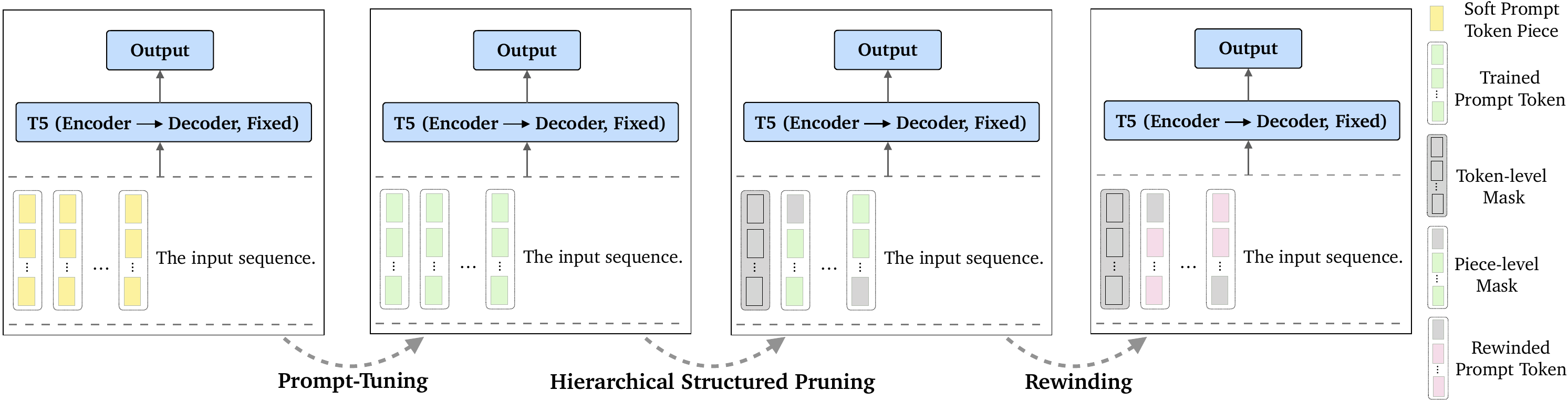}
  \caption{The illustration of our proposed \textsc{XPrompt} approach. \textsc{XPrompt} consists of three stages, namely \textit{Prompt-Tuning}, \textit{Hierarchical Structured Pruning} and \textit{Rewinding}. Among all the stages, the parameters of T5 are frozen - only the parameters of the prompts are tuned. The prompts trained in the previous stage are fed into the next stage as the initialization prompts. The change of color represents the process that the parameters of the prompts are tuned or pruned.}
  \label{Train_Mask_Rewind}
  \vspace{-4mm}
\end{figure*}

\subsection{Prompt Learning in NLP}
With the development of GPT-$3$ \citep{brown2020language}, prompt learning has drawn much attention in the NLP community \citep{liu2021pre, NNN2022DeltaTA}, which enables efficient learning by adding a number of \textit{prompt} tokens to the input. Prompt learning has been proven to be effective in various downstream tasks \citep{davison2019commonsense, gong2021prompt, radford2019language, wang2021transprompt, khashabi2020unifiedqa}. 
Recently, prompt has been extended from discrete tokens (tokens in the vocabularies) to continuous tokens (trainable embeddings), i.e., soft prompt \citep{li2021prefix, zhong2021factual, qin2021learning}. For example, \citep{lester2021power} proposes a parameter-efficient prompt tuning approach by only tuning soft prompts and fixing the entire parameters in PLM. Prompt tuning achieves great success and shows that it can reach the performance of fine-tuning with large PLM. However, there is still a large performance gap between prompt tuning and fine-tuning for models of moderate scales. More recently, \citep{vu2021spot} proposes a prompt-based transfer learning approach, \textsc{SPoT}, to improve the performance of prompt tuning, which learns a prompt on source tasks and then applied to initialize the target task's prompt. Most recently, \citep{he2022hyper} proposes HyperPrompt which uses the hypernetworks to generate hyper-prompts and obtains superior performance. However, it needs to tune all parameters and shows that only tuning task-conditioned parameters is not enough to achieve competitive results as full model fine-tuning for multi-task learning.

\subsection{Lottery Ticket Hypothesis}
The lottery ticket hypothesis \citep{frankle2018lottery} finds that an over-parameterized network contains a subnetwork that is initialized such that - when trained in isolation - it can match the test accuracy of the original network after training for at most the same number of iterations. The subnetwork is called lottery ticket. In NLP, the collection of lottery tickets is referred to as winning tickets in highly over-parametrized models, e.g., PLMs \citep{liang2021super}. Such winning tickets have demonstrated their abilities to transfer across tasks and datasets \citep{morcos2019one, yu2019playing, desai2019evaluating}. Recently, \citet{chen2021lottery} has shown the existence of the winning tickets in PLMs. \citet{liang2021super} observes that the generalization performance of the winning tickets can even exceed that of the full model.


\section{Preliminary}
Built upon the text-to-text approach of T$5$ \citep{raffel2020exploring}, prompt tuning formulates all tasks as text generation by prepending additional $l$ tunable soft prompt tokens to the input and only updating the parameters of the inserted soft prompt tokens. Specifically, given a series of $n$ input tokens $X=\{x_1, x_2, ... , x_n\}$,  T$5$ first generates the token embeddings $X_e \in \mathbb{R}^{n \times e}$, where $e$ is the dimension of the embedding space. It also generates soft prompt embeddings $P_e=\{p_1, p_2, ..., p_m\} \in \mathbb{R}^{m \times e}$, where $m$ is the length of the soft prompt. Then the soft prompts are prepended to the input sequence as $[P_e; X_e] \in \mathbb{R}^{(m+n) \times e}$. The goal of prompt tuning is to maximize the likelihood of the labels $Y$ by only optimizing over $P_e$:
\begin{equation}
  \mathop{\arg\max}_{P_e} \log p(Y| [P_e;X_e]) 
\end{equation} 

Prompt tuning becomes more effective as the model scale increases. However, there is still a significant performance gap between prompt tuning and fine-tuning especially for models of small and moderate scales. 
Our hypothesis is that not all soft prompt tokens contribute equally to the performance after training on the target task. There exist certain soft prompt tokens that may have negative impacts on the task. Therefore, combining the idea of the lottery ticket hypothesis, we propose \textsc{XPrompt} with hierarchical structured pruning to identify the optimal soft prompts and bridge the performance gap.

\section{\textsc{XPrompt}}
The overall process of \textsc{XPrompt} is illustrated in Figure~\ref{Train_Mask_Rewind}, which consists of three main stages: \textit{Prompt-Tuning}, \textit{Hierarchical Structured Pruning} and \textit{Rewinding}. Specifically, the prompt tuning learns an initial set of values for all soft prompt tokens on the target task. During the hierarchical structured pruning, token-level and piece-level pruning processes are repeatedly conducted to identify the optimal soft tokens and pieces at different compression ratios. Finally, a weight rewinding technique is applied to re-train the soft prompts.

\begin{figure}[t!]
  \centering
  \resizebox{0.42\textwidth}{!}{
  \includegraphics[width=\linewidth]{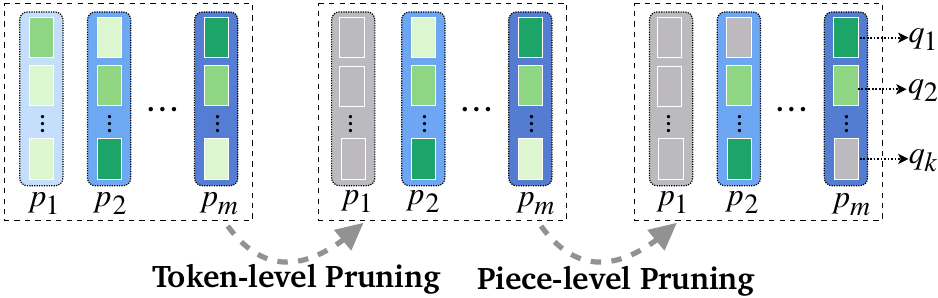}}
  \caption{The illustration of \textit{Hierarchical Structured Pruning}. Among them, the shade of the color indicates the level of the importance score, and the darker the color, the higher the importance score of the corresponding structure (token or piece).}
  \label{HierarchicalStructuredPruning}
  \vspace{-5mm}
\end{figure}

\subsection{Prompt Tuning}
Prompt tuning approaches prepend a number of soft prompt tokens to the input, and only tune soft prompts by fixing the entire parameters in PLM. Prompt tuning has been proven to be effective in various downstream tasks. In our prompt tuning stage, following previous work \citep{liang2021super}, we conduct a complete tuning on the target task to obtain the embeddings for all the soft prompt tokens. These trained soft prompts are used as initialization in the hierarchical structured pruning.

\subsection{Hierarchical Structured Pruning}
Hierarchical structured pruning is designed to separate negative prompt tokens from the trained prompt tokens, and identify an optimal set of soft prompts. 
The approach is illustrated in Figure~\ref{HierarchicalStructuredPruning}. The token-level pruning is first used to identify negative prompt tokens, however, the rest prompt tokens may still contain negative pieces. Thus, the piece-level pruning is then applied to identify more fine-grained negative prompt pieces within each prompt token. Token-level and piece-level pruning together play a better trade-off between effectiveness and efficiency.

\subsubsection{Token-level Pruning}

To identify negative prompt tokens in the trained prompt tokens, we associate mask variable $\gamma_i$ to each soft prompt token vector $p_i$:
\begin{equation}
  \hat{P_e} = \gamma \cdot  P_e
\end{equation} 
where $\gamma = \{\gamma_1, \gamma_2, ..., \gamma_m \}, \gamma_i \in \{0, 1\}$, and a 0 value indicates that the corresponding soft prompt token is pruned. 


We then calculate the importance score \citep{michel2019sixteen} of each token to distinguish the negative prompt tokens from the other ones. The importance score is defined as the expected sensitivity of the model outputs to the mask variables. Formally, the importance score $I_{p_i}$ of each soft prompt token $p_i$ is calculated as:
\begin{equation}
 I_{p_i} =
 \mathbb{E}_{x \sim \mathcal{D}_x}\mid \frac{\partial  \mathcal{L}(x) }{\partial \gamma_i } \mid 
\end{equation} 
where $\mathcal{L}$ is the loss function and $\mathcal{D}_x$ is the training data distribution. 

Essentially, the importance score of each soft prompt token indicates its individual contribution to the model performance. A low importance score means that the corresponding soft prompt token has a small or even negative contribution to the model. In other words, such a soft prompt token contains negligible prompt information for generating the outputs. On the contrary, a large importance score implies a major contribution with more meaningful prompt information. Therefore, the prompt tokens with low importance scores are most likely negative prompt tokens, which are pruned during the token-level pruning stage.

\subsubsection{Piece-level Pruning}
Token-level pruning finds the most important soft prompt tokens. However, it may not be sufficient as there are still fine-grained negative prompt pieces remaining in the embedding of each soft prompt token. Different pieces of the embedding may lead to different effects on downstream tasks. Therefore, we further conduct piece-level pruning to eliminate the negative prompt pieces within each token. In particular, we divide the embedding vector of each soft prompt token ${p_i}_e$ into k pieces with equal scale, $q_e = \{ {q_1}_e, {q_2}_e, ..., {q_k}_e \}$, and treat each piece as an independent unit that can be optimized with gradient updates. 
Mask variable $\zeta_i$ is associated with each piece in the soft prompt token to identify the negative prompt pieces:
\begin{equation}
  \hat{q_e} = \zeta \cdot  q_e
\end{equation} 
where $\zeta = \{ \zeta_1, \zeta_2, ..., \zeta_k \}, \zeta_i \in \{0,1\}$, and 0 value indicates that the corresponding piece is pruned. 

We then calculate the importance score $I_{q_i}$ of each piece for every prompt token embedding to prune the low-importance pieces:
\begin{equation}
 I_{q_i} =
 \mathbb{E}_{x \sim \mathcal{D}_x}\mid \frac{\partial  \mathcal{L}(x) }{\partial \zeta_i } \mid 
\end{equation}

Similar to the token-level importance score, a low piece-level importance score indicates that the piece has a small or even negative contribution towards the model performance. Such low-importance pieces contain limited information for generating the outputs. We repeatedly conduct both token-level and piece-level pruning to obtain the sub-prompt tokens and pieces at different compression ratios.

\subsection{Rewinding}
The lottery ticket hypothesis (LTH) \citep{frankle2018lottery} states that sparse subnetworks (the unpruned prompts) can be trained in isolation to the same accuracy as the original network (all prompts), and proposes training to pruning and then rewinding the unpruned weights.
Following the idea in LTH, we adopt the weight rewinding technique \citep{renda2019comparing} to re-train the soft prompts after the two-level hierarchical structured pruning. Specifically, we reset the parameters of the selected optimal soft prompts using their values after the prompt tuning stage. The other soft prompts are pruned by setting the corresponding mask variables to $0$. Finally, we re-train the soft prompts using the original learning strategies in prompt tuning.

\section{Experiments}

\subsection{Datasets}
To cover broad and diverse NLP tasks in our experiments, we evaluate our method on various datasets of SuperGLUE benchmark \citep{wang2019superglue} in both high-resource and low-resource scenarios. Due to restricted test access for SuperGLUE, following previous works \citep{lester2021power, ding2021openprompt}, we tune the prompt model on the training set for a fixed number of steps and report results on the validation set using the best checkpoint. The detailed description, statistics and metrics of SuperGLUE tasks are provided in Table~\ref{SuperGLUE} of Appendix  ~\ref{section:SuperGLUE_Statistics_and_Metrics}. The soft prompt templates and generation verbalizers are provided in Table~\ref{SoftTemplate} of Appendix ~\ref{section:SuperGLUE_Statistics_and_Metrics}.

\subsection{Baselines}
\paragraph{Fine-Tuning} We compare with the standard fine-tuning approach \citep{raffel2020exploring, aribandi2021ext5} of T$5$, where all the pre-trained parameters are fine-tuned on each target task separately.

\paragraph{Prompt-Tuning} The vanilla prompt tuning approach of \citep{lester2021power} showed that prompt tuning is a competitive technique for adapting frozen PLMs to downstream tasks. 


\paragraph{P-Tuning}\citep{liu2021gpt} is a prompt-based method that uses the masked PLM to convert the target task into a cloze problem. It employs soft-prompting techniques to optimize prompts in the continuous space.
We also compare with its second version P-TuningV$2$ \citep{liu2021p}.

\paragraph{Prefix-Tuning}\citep{li2021prefix} is a lightweight alternative to fine-tuning for natural language generation tasks, which only optimizes a small continuous task-specific vector (called prefix). Prefix-Tuning prepends the prefix to inputs of every transformer layer independently.

\subsection{Implementation}
Our method is implemented with the OpenPrompt library \citep{ding2021openprompt}, which is a unified and extensible toolkit for prompt learning. We translate each SuperGLUE dataset into a text-to-text format following \citep{raffel2020exploring}, except that we omit the task names prepend to inputs indicating which SuperGLUE task an example belongs to.

Our \textsc{XPrompt} is built on top of the pre-trained T$5$ checkpoints of three scales: \texttt{Large}, \texttt{XL}, \texttt{XXL} with $770$M, $3$B and $11$B parameters, respectively. Following previous studies \citep{lester2021power, ding2021openprompt}, we train our prompts for $100$ epochs with a constant learning rate of $0.3$ and a batch size of $16$. \citep{lester2021power} shows that an increase beyond $20$ tokens only yields marginal gains, so throughout our experiments, we set the default number of prompt tokens to $20$ to control the number of trainable parameters and use sampled vocabulary to initialize the prompt parameters. The number of pieces in each token is set to $16$. The pruning frequencies are linearly searched from \{10\%, 20\%, 30\%, 40\%, 50\%, 60\%, 70\%, 80\%, 90\%\}. The weight rewinding is applied only once to re-train the pruned soft prompts. The best checkpoints are selected via early stopping on the development set.
The models are trained using the Adafactor \citep{shazeer2018adafactor} optimizer with weight decay $1e$-$5$.

\begin{table*}[ht]
\centering
\resizebox{1.0\textwidth}{!}{
\begin{tabular}{cc|cccccccc}
\toprule
\multicolumn{2}{c|}{\textbf{Model}} & \textbf{WiC}  & \textbf{WSC}  & \textbf{CB}  & \textbf{COPA}  & \textbf{RTE}   & \textbf{Boolq} & \textbf{MultiRC}     & \textbf{Average}   \\  
&   &Acc &Acc &Acc &Acc &Acc &Acc &F1$_{a}$  &Score \\
\hline
\multicolumn{1}{c|}{\multirow{4}{*}{\begin{tabular}[c]{@{}c@{}}\textbf{T5-} \\
\textbf{Large}\\ 770M\end{tabular}}} 
& Fine-Tuning$^\ast$ & 73.50 & 88.50 & 94.30 & 72.0 & 90.60 & 88.30  &85.40     & 84.65    \\ \cline{3-10}
\multicolumn{1}{c|}{}                          
& P-Tuning   & 70.37  & 64.42 & 92.85 & 76.0 & 79.78 & 83.02  &79.96    & 78.06     \\
\multicolumn{1}{c|}{}                          
& Prefix-Tuning   &62.50   &64.46  &78.78  &-  &55.70  &65.17   &60.19    &64.46      \\
\multicolumn{1}{c|}{}                          
& Prompt-Tuning & 72.25 & 68.26 & 82.14 & 76.0 & 85.19 & 83.02  &79.86  & 78.10     \\
\multicolumn{1}{c|}{}  & \textbf{\textsc{XPrompt}} & \textbf{73.51}\small{\textbf{{\textcolor{blue}{$\uparrow$1.26}}}} &  \textbf{70.39}\small{\textbf{{\textcolor{blue}{$\uparrow$2.13}}}} &  \textbf{91.07}\small{\textbf{{\textcolor{blue}{$\uparrow$8.93}}}} & \textbf{82.0}\small{\textbf{{\textcolor{blue}{$\uparrow$6.0}}}} & \textbf{87.72}\small{\textbf{{\textcolor{blue}{$\uparrow$2.53}}}} & \textbf{83.82}\small{\textbf{{\textcolor{blue}{$\uparrow$0.8}}}}  &
\textbf{81.02}\small{\textbf{{\textcolor{blue}{$\uparrow$1.16}}}}  & \textbf{81.36}\small{\textbf{{\textcolor{blue}{$\uparrow$3.26}}}}  \\ \hline
\multicolumn{1}{c|}{\multirow{4}{*}{\begin{tabular}[c]{@{}c@{}}\textbf{T5-XL}\\ 3B \end{tabular}}}    
& Fine-Tuning$^\ast$ & 74.30 & 95.20  & 92.00 & 96.0 & 91.70 & 89.60  &88.20   & 89.57 \\ \cline{3-10}
\multicolumn{1}{c|}{}                          
& P-Tuning  & 72.54  & 81.73  & 91.07   & 73.0    & 89.53   & 84.54  &85.45 & 82.55 \\
\multicolumn{1}{c|}{}                          
& Prompt-Tuning  & 74.29  & 86.53 & 91.07  & 91.0  & 89.16  & 87.58   &84.89   & 86.36  \\
\multicolumn{1}{c|}{}    & \textbf{\textsc{XPrompt}} & \textbf{76.95}\small{\textbf{{\textcolor{blue}{$\uparrow$2.66}}}} & \textbf{91.34}\small{\textbf{{\textcolor{blue}{$\uparrow$4.84}}}} & \textbf{92.85}\small{\textbf{{\textcolor{blue}{$\uparrow$1.78}}}}  & \textbf{95.0}\small{\textbf{{\textcolor{blue}{$\uparrow$4.0}}}} & \textbf{92.79}\small{\textbf{{\textcolor{blue}{$\uparrow$3.63}}}} & \textbf{89.00}\small{\textbf{{\textcolor{blue}{$\uparrow$1.42}}}} &  
\textbf{87.34}\small{\textbf{{\textcolor{blue}{$\uparrow$2.45}}}}  & \textbf{89.32}\small{\textbf{{\textcolor{blue}{$\uparrow$2.96}}}}   \\  \hline
\multicolumn{1}{c|}{\multirow{4}{*}{\begin{tabular}[c]{@{}c@{}}\textbf{T5-XXL}\\ 11B \end{tabular}}}   
& Fine-Tuning$^\ast$ & 78.50 & 95.20 & 100.00 & 99.0 & 92.10 & 90.40  &88.60  & 91.97 \\ \cline{3-10}
\multicolumn{1}{c|}{}          
& P-Tuning & 76.80  & 94.23  & 92.85 & 93.0  & 89.80   & 86.98   &87.56   & 88.75 \\
\multicolumn{1}{c|}{}                         
& Prompt-Tuning   & 76.10  & 96.15  & 96.42  & 98.0 & 91.69  & 89.08   &87.90  & 90.76 \\
\multicolumn{1}{c|}{}   & \textbf{\textsc{XPrompt}} &  \textbf{77.69}\small{\textbf{{\textcolor{blue}{$\uparrow$1.59}}}} & \textbf{97.11}\small{\textbf{{\textcolor{blue}{$\uparrow$0.96}}}} & \textbf{100.00}\small{\textbf{{\textcolor{blue}{$\uparrow$3.58}}}} & \textbf{99.0}\small{\textbf{{\textcolor{blue}{$\uparrow$1.0}}}} & \textbf{94.94}\small{\textbf{{\textcolor{blue}{$\uparrow$3.25}}}} & \textbf{90.87}\small{\textbf{{\textcolor{blue}{$\uparrow$1.79}}}} & 
\textbf{88.90}\small{\textbf{{\textcolor{blue}{$\uparrow$1.0}}}} &  \textbf{92.64}\small{\textbf{{\textcolor{blue}{$\uparrow$1.88}}}} \\ 
\bottomrule
\end{tabular}}
\caption{Main experimental results (\%) on seven SuperGLUE tasks. Our method and better results are in bold (the larger, the better). The small number next to each score indicates performance improvement (\textcolor{blue}{$\uparrow$}) compared with the vanilla Prompt-Tuning. Methods with `$^\ast$' indicate the results reported in ~\citet{aribandi2021ext5}. We only present the results of Prefix-Tuning on T$5$-Large, since it can diverge with larger models~\citep{NNN2022DeltaTA}. The `-' results in Prefix-Tuning indicate diverged results in the corresponding task.}
 \label{mainSuperGLUEresults}
 \vspace{-4mm}
\end{table*}

\section{Results}


\subsection{Results on High-resource Scenarios}

\textbf{\textsc{XPrompt} significantly improves the performance of prompt tuning and helps close the gap with fine-tuning across all model scales.} Table \ref{mainSuperGLUEresults} and Table~\ref{PtuningV2results} (in the appendix) present the main results on SuperGLUE. We compare \textsc{XPrompt} with strong prompt learning baselines, including Prompt-Tuning, Prefix-Tuning, P-Tuning and P-TuningV$2$ for different PLMs and model scales.
It can be seen that \textsc{XPrompt} outperforms vanilla Prompt-Tuning by a large margin across all tasks and model scales. For instance, \textsc{XPrompt} yields an improvement of $3.26$ \%, $2.96$ \%, and $1.88$ \% in terms of average score on T$5$-Large, T$5$-XL, and T$5$-XXL, respectively. We also observe that the performance of Prompt-Tuning and P-Tuning are comparable at the same model scale. Moreover, P-TuningV$2$ outperforms Prompt-Tuning and P-Tuning on CB, RTE, and Boolq. However, \textsc{XPrompt} achieves more predominant performances than P-TuningV$2$ at similar model scales, demonstrating its effectiveness. It is worth noting that Prefix-Tuning is less performable on most NLU tasks, since it is designed for natural language generation (NLG) tasks.


It is clear from Table \ref{mainSuperGLUEresults} that \textsc{XPrompt} enables prompt tuning to match the fine-tuning performance on all tasks with T$5$-XL, and even exceeds fine-tuning performance on most tasks at the T$5$-XXL scale. For example, \textsc{XPrompt} achieves the best average score of $89.32$\% with T$5$-XL, leaving only $0.25$\% gap to fine-tuning. It is worth mentioning that \textsc{XPrompt} significantly outperforms fine-tuning on WiC, CB and RTE with T$5$-XL, as well as COPA and WiC with T$5$-Large. Especially for T$5$-XXL, \textsc{XPrompt} achieves the best score of $97.11$\%, $100.00$\%, $94.94$\%, $90.87$\% and $88.90$\% on WSC, CB, RTE, Boolq, MultiRC respectively, leading to +$1.91$\%, +$0.0$\%, +$2.84$\%, +$0.47$\%, +$0.30$\% improvements over fine-tuning. 
We also observe that there are certain gaps between prompt tuning and fine-tuning, especially for small and moderate scale models (see Figure \ref{SuperGLUE_Line}). However, our \textsc{XPrompt} narrows down the gap significantly across all model scales, demonstrating that it learns efficient and informative soft prompts which empower downstream tasks effectively.

\begin{table}[t]
 \centering
\resizebox{0.36\textwidth}{!}{
\begin{tabular}{cccc}
\toprule
\textbf{Model}   & \textbf{Boolq} & \textbf{WiC} & \textbf{RTE}  \\ \hline
\multicolumn{1}{c|}{P-Tuning} & 64.99  & 54.23 & 57.40     \\
\multicolumn{1}{c|}{GPT-3 XL1.3B$^{\ddag}$}  & 64.10  & 53.00  & 50.90   \\
\multicolumn{1}{c|}{GPT-3 2.7B$^{\ddag}$}    & \textbf{70.30}  & 51.60  & 56.30   \\
\multicolumn{1}{c|}{PromptTuning} & 69.81    & 60.81  & 66.08            \\
\multicolumn{1}{c|}{\textbf{\textsc{XPrompt}}} & 70.23   &\textbf{62.85}  &\textbf{67.87}   \\ 
\bottomrule
\end{tabular}}
\caption{The few-shot ($32$ samples) results (Acc, \%) on three SuperGLUE tasks for the T$5$-XL model with $20$ soft prompt tokens. Methods with `$^\ddag$' indicate results reported in \citet{schick2021s}. \textsc{XPrompt} is better than vanilla Prompt-Tuning and P-Tuning in low resource scenarios.}
 \label{fewshotResults}
 \vspace{-5.5mm}
\end{table}

\subsection{Results on Low-resource Scenarios}
\textbf{\textsc{XPrompt} performs much better in low resource scenarios.} Since prompt learning is surprisingly effective in low-resource regime \citep{schick2021s}, we also explore the effect of \textsc{XPrompt} in low-resource scenarios. Following the setting used in~\citep{schick2021s}, we randomly select $32$ examples as the new training set for each task using a fixed random seed. We tune the prompt model on the $32$-shot training set and directly report the full dev set results using the best checkpoint.

As demonstrated in Table \ref{fewshotResults}, our \textsc{XPrompt} further improves the performance of prompt tuning and outperforms the baseline models at the same scale on Boolq, WiC, and RTE. For example, \textsc{XPrompt} achieves the best score of $62.85$\% on WiC, +$2.04$\% improvement over Prompt-Tuning. These few-shot results suggest that although overfitting is severe especially when training with limited data, \textsc{XPrompt} consistently lifts the performance of prompt tuning.

\section{Analysis and Discussion}
To better understand the effectiveness of the \textsc{XPrompt} and explore the impact of various factors in \textsc{XPrompt}, we further conduct a series of ablation studies and analysis.

\subsection{Do Positive Prompts and Negative Prompts Exist?}
\textbf{We identify both positive and negative prompts through hierarchical structured pruning.} For positive prompts, the first evidence is the large performance improvement of \textsc{XPrompt} over vanilla prompt tuning across all tasks and model scales, which shows the effectiveness of these positive prompts. Another evidence is the high sparsities of pruning. Figure~\ref{visual_importance_scores_piece_original} and Figure~\ref{visual_importance_scores_piece_masked} in Appendix~\ref{section:importance_scores_masked_visua} show the original and pruned gradient saliency maps \citep{simonyan2014deep} of the importance scores on WSC task, i.e., the gray elements in Figure~\ref{visual_importance_scores_piece_masked} indicate that the prompt tokens or pieces are pruned due to low importance scores, and the remaining parts are the winning tickets. The performance of \textsc{XPrompt} with $15$\% positive sub-prompts is $4.8$\% higher than the full prompt tuning.

\begin{figure}[t]
  \centering
  \resizebox{0.4\textwidth}{!}{
  \includegraphics[width=\linewidth]{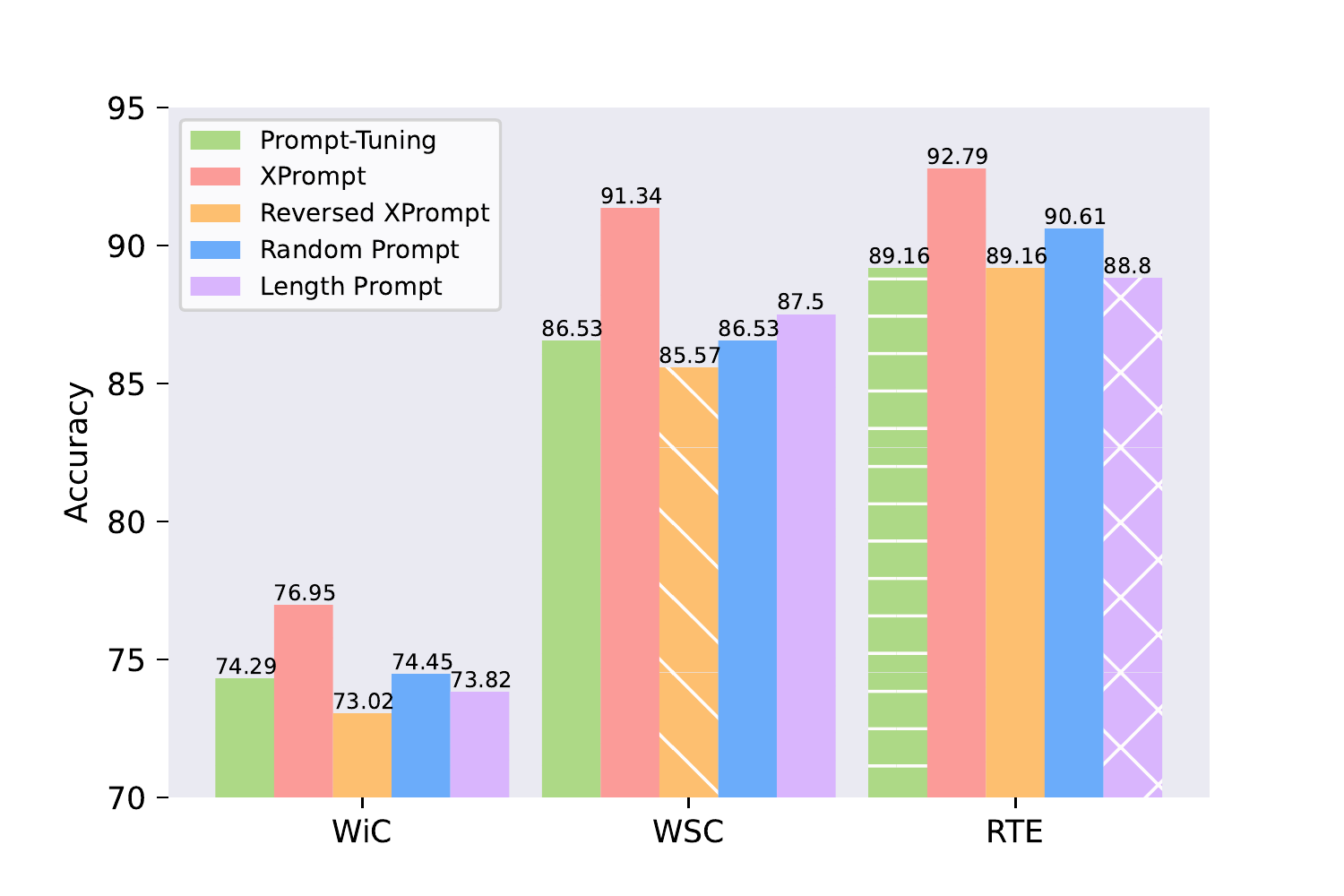}}
  \caption{The performance of Prompt-Tuning, \textsc{XPrompt}, Reversed \textsc{XPrompt}, Random Prompts and Length Prompt comparison with T$5$-XL model on three tasks. Among them, Reversed \textsc{XPrompt} denotes the masked sub-prompt, Random Prompt denotes the randomly masked sub-prompt, and Length Prompt denotes the reserved prompt whose prompt length is the same as \textsc{XPrompt}.}
  \label{original_tickets_masked}  
  \vspace{-4mm}
\end{figure}

\textbf{The negative prompts perform worse than Prompt Tuning and \textsc{XPrompt}}. 
To further investigate the existence and effect of negative prompts, we conduct another experiment to compare prompt tuning performances with different configurations. Specifically, in addition to the vanilla Prompt-Tuning (using all prompts) and our \textsc{XPrompt}, we introduce three variations - Reversed \textsc{XPrompt}, Random Prompt and Length Prompt.   
The Reversed \textsc{XPrompt} reverses the masked sub-prompt structures in \textsc{XPrompt}, which essentially uses all the low score prompt tokens and pieces. For Random Prompt, we mask tokens and pieces randomly at the rewind stage. The Length Prompt retrains prompt tuning with the same prompt length of the resulting \textsc{XPrompt}. The comparison results are shown in Figure~\ref{original_tickets_masked}. It can be seen that our \textsc{XPrompt} achieves the best performance among them. We also observe that the Reversed \textsc{XPrompt} performs significantly worse than all other prompt tuning variants, including Random Prompt and Length Prompt. This observation is consistent with our expectation and further validates the existence of the negative prompts. It is worth noting that the Length Prompt performs worse than Random Prompt and Prompt Tuning on average, indicating the effectiveness of our hierarchical structured pruning. The distribution of the importance scores of the prompt tokens is shown in Figure~\ref{PromptTokensImportanceScore} in the appendix.

\begin{table}[ht]
\resizebox{0.50\textwidth}{!}{
\begin{tabular}{cccccccc}
\toprule
\textbf{Model} & \textbf{WiC} & \textbf{WSC} & \textbf{CB} & \textbf{COPA} & \textbf{RTE} & \textbf{Boolq} & \textbf{MultiRC} \\ \hline
Fine-Tuning &$3\times{10}^{9}$ &$3\times{10}^{9}$ &$3\times{10}^{9}$ &$3\times{10}^{9}$      &$3\times{10}^{9}$ &$3\times{10}^{9}$ &$3\times{10}^{9}$   \\
Prompt-Tuning  &40960     &40960     &40960    &40960      &40960     &40960   &40960    \\
\textbf{\textsc{XPrompt}} & 2560 & 6144   & 2560    &15232      &512     &29184  &27648      \\
\textbf{Percentage} &6.25\%     &15\%     &6.25\%    &37.18\%      &1.25\%     & 71.25\%  &67.5\%     \\ 
\bottomrule
\end{tabular}}
\caption{The number of tunable parameters comparison for T$5$-XL model with $20$ prompt tokens. The percentage means the number of tunable parameters in \textsc{XPrompt} compared to Prompt-Tuning. }
 \label{parameters}
 \vspace{-4mm}
\end{table}

\subsection{Parameter Efficiency}
\textbf{\textsc{XPrompt} is more parameter-efficient than Prompt-Tuning.} The number of tunable parameters comparison is shown in Table \ref{parameters}. Clearly, Prompt-Tuning is already parameter-efficient, which only needs to tune $0.0014$\% parameters compared to full model fine-tuning. However, \textsc{XPrompt} further reduces the tunable parameters in Prompt-Tuning through hierarchical structured pruning. For instance, \textsc{XPrompt} only tunes $15$\% and $37.18$\% parameters compared to Prompt-Tuning.

\begin{table}[ht]
\centering
\resizebox{0.42\textwidth}{!}{
\begin{tabular}{cc|cccc}
\toprule
\multicolumn{2}{c|}{\textbf{Model}} 
& \textbf{WSC} & \textbf{CB} & \textbf{COPA} & \textbf{RTE} \\ \hline
\multicolumn{1}{c|}{\multirow{4}{*}{\textbf{\begin{tabular}[c]{@{}c@{}}T5-\\ Large\end{tabular}}}} & Prompt-Tuning  & 68.26  & 82.14   & 76.0   & 85.19        \\
\multicolumn{1}{c|}{} & Token-level & 70.19  & 91.07   & 80.0   & 86.28        \\
\multicolumn{1}{c|}{} & Piece-level & 69.23  & 89.28   & 79.0   & 86.64        \\
\multicolumn{1}{c|}{} &\textbf{\textsc{XPrompt}} &\textbf{70.39} & \textbf{91.07}  & \textbf{82.0}  &\textbf{87.72}  \\ \hline
\multicolumn{1}{c|}{\multirow{4}{*}{\textbf{T5-XL}}} 
& Prompt-Tuning & 86.53 & 91.07 & 91.0  & 89.16 \\
\multicolumn{1}{c|}{}  & Token-level & 89.42 & 92.85 & 93.0  & 92.41        \\
\multicolumn{1}{c|}{}  & Piece-level & 90.38 & 92.85 & 93.0  & 91.33        \\
\multicolumn{1}{c|}{}  & \textbf{\textsc{XPrompt}} &\textbf{91.34} & \textbf{92.85} &\textbf{95.0} &\textbf{92.79}        \\
\bottomrule
\end{tabular}}
\caption{The results of different pruning levels on four SuperGLUE tasks using T$5$-Large and T$5$-XL models.}
 \label{TwoLevel_SuperGLUE_results}
 \vspace{-4mm}
\end{table}

\subsection{Granularity of Pruning}
\textbf{Token-level pruning and fine-grained piece-level pruning are both important.} To further investigate the effects of the two-level pruning, we conduct extensive ablation experiments on four SuperGLUE tasks, whose results are included in Table~\ref{TwoLevel_SuperGLUE_results}. In general, both two levels of structured pruning outperform vanilla Prompt-Tuning, demonstrating the effectiveness of both token-level and piece-level pruning. The results also show the existence of sub-prompt structures in trained prompts that can be further optimized. Obviously, \textsc{XPrompt} outperforms individual one level pruning, which suggests the combination of the two levels of structured pruning further benefits the training of the soft prompts for downstream tasks. 

\begin{table}[ht]
  \centering
\resizebox{0.42\textwidth}{!}{
\begin{tabular}{cccccc}
\toprule
\textbf{Length} &\textbf{Model} &\textbf{WSC} &\textbf{CB} &\textbf{COPA} &\textbf{RTE}  \\ \hline
\multicolumn{1}{c|}{\multirow{2}{*}{\textbf{10}}}  & \multicolumn{1}{c|}{Prompt-Tuning}   & 82.69       & 87.50       & 87.0      & 88.44      \\
\multicolumn{1}{c|}{}                     & \multicolumn{1}{c|}{\textbf{\textsc{XPrompt}}} & 87.50      & 91.07      & 93.0      & 90.61      \\ \hline
\multicolumn{1}{c|}{\multirow{2}{*}{\textbf{20}}}  & \multicolumn{1}{c|}{Prompt-Tuning}   & 86.53 & 91.07 & 91.0 & 89.16 \\
\multicolumn{1}{c|}{}                     & \multicolumn{1}{c|}{\textbf{\textsc{XPrompt}}} & 91.34 & 92.85 & 95.0 & 92.79 \\ \hline
\multicolumn{1}{c|}{\multirow{2}{*}{\textbf{100}}} & \multicolumn{1}{c|}{Prompt-Tuning}   & 89.42      &91.07       & 90.0      & 89.16      \\
\multicolumn{1}{c|}{}                     & \multicolumn{1}{c|}{\textbf{\textsc{XPrompt}}} & 91.94       &92.85       &94.0      &92.79       \\ 
\bottomrule
\end{tabular}}
\caption{The results of different prompt lengths on four SuperGLUE tasks using the T$5$-XL model.}
 \label{PromptLength_SuperGLUE_results}
 \vspace{-4mm}
\end{table}

\subsection{Prompt Length}
\textbf{Increasing prompt length (beyond 20) only yields marginal gains for \textsc{XPrompt}.} To explore the effect of prompt length on \textsc{XPrompt}, we train \textsc{XPrompt} for the T$5$-XL model with different prompt lengths in \{$10$, $20$, $100$\}. The results are reported in Table \ref{PromptLength_SuperGLUE_results}. From these results we can see that although prompt length plays an importance role for \textsc{XPrompt} and Prompt-Tuning,
the improvements are limited when increasing the prompt length to beyond 20 tokens. This observation is consistent with the findings in \citep{lester2021power}, and that is why we set the number of prompt tokens to 20 in all our experiments.

\subsection{Prompt Initialization and Transfer}
Motivated by the soft prompts transfer approach (\textsc{SPoT}) \citep{vu2021spot}, to explore the effect of task transfer and different prompt initialization methods, we introduce a \textsc{XPrompt} based transfer learning approach - \textsc{XPrompt} Transfer. It first trains the prompts through \textsc{XPrompt} on the source task and then uses the learned prompts to initialize the prompts on the target task. More details are provided in Appendix~\ref{section:XPrompt_Transfer}.
\begin{table}[t]
\centering
\resizebox{0.40\textwidth}{!}{
\begin{tabular}{cc|cc}
\toprule
\multicolumn{2}{c|}{Initialization Methods}                                                                                           & \textbf{WSC}   & \textbf{COPA} \\ \hline
\multicolumn{2}{c|}{Prompt-Tuning(SampledVocab)}                                                                        & 86.53 & 91.0 \\ \hline
\multicolumn{1}{c|}{\multirow{2}{*}{\begin{tabular}[c]{@{}c@{}}XPrompt\\ Initialization\end{tabular}}} & RandomUniform & 88.61 & 93.0 \\
\multicolumn{1}{c|}{}                                                                                  & SampledVocab  & 91.34 & 95.0 \\ \bottomrule
\end{tabular}}
\caption{The results of different prompt initialization methods for \textsc{XPrompt} on two SuperGLUE tasks using T$5$-XL model. }
\label{initializationResults}
\vspace{-3mm}
\end{table}

\begin{table}[t]
\centering
\resizebox{0.36\textwidth}{!}{
\begin{tabular}{c|ccc}
\toprule
TransferMethod  & \textbf{WSC} &$\Leftrightarrow$ & \textbf{COPA} \\ \hline
TaskTransfer    & 86.53        &  & 92.0          \\
XPromptTransfer$_o$ & 86.93        &  & 95.0          \\
XPromptTransfer & 91.40        &  & 98.0          \\ \bottomrule
\end{tabular}}
\caption{The results of \textsc{XPrompt} Transfer on two SuperGLUE tasks using T$5$-XL model. \textsc{XPrompt}Transfer$_o$ only uses the resulting prompts of the source task through \textsc{XPrompt} to initialize the prompts of the target task, without the rewinding phase.}
\label{PromptTransfer_results}
\vspace{-4mm}
\end{table}

\textbf{Prompt initialization plays an important role in \textsc{XPrompt}, and \textsc{XPrompt} Transfer can lead to performance gains.} We compare two sample initialization methods for \textsc{XPrompt}, including random uniform and sampled vocabulary, the results are shown in Table~\ref{initializationResults}. We observe that sampled vocabulary performs best, and our \textsc{XPrompt} can also lead to performance gains for the random uniform initialization. Furthermore, we compare our \textsc{XPrompt} Transfer with the TaskTransfer, which only uses the resulting prompts of the source task to initialize the prompts of the target task, the results are shown in Table~\ref{PromptTransfer_results}. We can see that \textsc{XPrompt} Transfer without rewinding stage outperforms the TaskTransfer, resulting in large performance gains through the pruning and rewinding. These results further validate our hypothesis and the effect of our \textsc{XPrompt} Transfer.

\section{Conclusions}
This paper aims to close the large performance gap between prompt tuning and fine-tuning, especially for models of small and moderate scales. By exploring the lottery ticket hypothesis in the context of prompt tuning, we have proposed a novel hierarchical structured pruning approach, namely \textsc{XPrompt}, to separate the positive prompts from the negative ones at both token-level and piece-level. Extensive experimental results have demonstrated that \textsc{XPrompt} yields  a more parameter-efficient prompt at an extremely small scale, yet with a competitive performance in effectiveness. Taken as a whole, our work sheds light on the development of more efficient and effective prompt-based learning approaches.

\section*{Limitations}
Eliminating negative prompt tokens at different granularity levels through hierarchical structured pruning requires rewinding the pruned model at different compression ratios. Therefore, a key question is left under-explored: how to find the optimal compression ratio without trial training, which can automate the training process and improve the efficiency. Moreover, there are other scenarios in prompt tuning that we plan to further investigate, including the multi-task learning scenario \citep{he2022hyper}, out-of-domain (domain shift) scenario \citep{lester2021power}, and prompt ensembling scenario \citep{lester2021power}. We leave these for future research.

\section*{Acknowledgments}
This research was supported in part by Natural Science Foundation of Beijing (grant number:  4222036) and Huawei Technologies (grant number: TC20201228005).

\bibliography{emnlp2022}

\begin{thebibliography}{46}
\expandafter\ifx\csname natexlab\endcsname\relax\def\natexlab#1{#1}\fi

\bibitem[{Aribandi et~al.(2021)Aribandi, Tay, Schuster, Rao, Zheng, Mehta,
  Zhuang, Tran, Bahri, Ni, Gupta, Hui, Ruder, and Metzler}]{aribandi2021ext5}
Vamsi Aribandi, Yi~Tay, Tal Schuster, Jinfeng Rao, Huaixiu~Steven Zheng,
  Sanket~Vaibhav Mehta, Honglei Zhuang, Vinh~Q. Tran, Dara Bahri, Jianmo Ni,
  Jai~Prakash Gupta, Kai Hui, Sebastian Ruder, and Donald Metzler. 2021.
\newblock \href {http://arxiv.org/abs/2111.10952} {Ext5: Towards extreme
  multi-task scaling for transfer learning}.
\newblock \emph{CoRR}, abs/2111.10952.

\bibitem[{Brown et~al.(2020)Brown, Mann, Ryder, Subbiah, Kaplan, Dhariwal,
  Neelakantan, Shyam, Sastry, Askell, Agarwal, Herbert{-}Voss, Krueger,
  Henighan, Child, Ramesh, Ziegler, Wu, Winter, Hesse, Chen, Sigler, Litwin,
  Gray, Chess, Clark, Berner, McCandlish, Radford, Sutskever, and
  Amodei}]{brown2020language}
Tom~B. Brown, Benjamin Mann, Nick Ryder, Melanie Subbiah, Jared Kaplan,
  Prafulla Dhariwal, Arvind Neelakantan, Pranav Shyam, Girish Sastry, Amanda
  Askell, Sandhini Agarwal, Ariel Herbert{-}Voss, Gretchen Krueger, Tom
  Henighan, Rewon Child, Aditya Ramesh, Daniel~M. Ziegler, Jeffrey Wu, Clemens
  Winter, Christopher Hesse, Mark Chen, Eric Sigler, Mateusz Litwin, Scott
  Gray, Benjamin Chess, Jack Clark, Christopher Berner, Sam McCandlish, Alec
  Radford, Ilya Sutskever, and Dario Amodei. 2020.
\newblock \href
  {https://proceedings.neurips.cc/paper/2020/hash/1457c0d6bfcb4967418bfb8ac142f64a-Abstract.html}
  {Language models are few-shot learners}.
\newblock In \emph{Advances in Neural Information Processing Systems 33: Annual
  Conference on Neural Information Processing Systems 2020, NeurIPS 2020,
  December 6-12, 2020, virtual}.

\bibitem[{Chen et~al.(2021)Chen, Frankle, Chang, Liu, Zhang, Carbin, and
  Wang}]{chen2021lottery}
Tianlong Chen, Jonathan Frankle, Shiyu Chang, Sijia Liu, Yang Zhang, Michael
  Carbin, and Zhangyang Wang. 2021.
\newblock \href
  {https://openaccess.thecvf.com/content/CVPR2021/html/Chen\_The\_Lottery\_Tickets\_Hypothesis\_for\_Supervised\_and\_Self-Supervised\_Pre-Training\_in\_CVPR\_2021\_paper.html}
  {The lottery tickets hypothesis for supervised and self-supervised
  pre-training in computer vision models}.
\newblock In \emph{{IEEE} Conference on Computer Vision and Pattern
  Recognition, {CVPR} 2021, virtual, June 19-25, 2021}, pages 16306--16316.
  Computer Vision Foundation / {IEEE}.

\bibitem[{Clark et~al.(2019)Clark, Lee, Chang, Kwiatkowski, Collins, and
  Toutanova}]{clark2019boolq}
Christopher Clark, Kenton Lee, Ming{-}Wei Chang, Tom Kwiatkowski, Michael
  Collins, and Kristina Toutanova. 2019.
\newblock \href {https://doi.org/10.18653/v1/n19-1300} {Boolq: Exploring the
  surprising difficulty of natural yes/no questions}.
\newblock In \emph{Proceedings of the 2019 Conference of the North American
  Chapter of the Association for Computational Linguistics: Human Language
  Technologies, {NAACL-HLT} 2019, Minneapolis, MN, USA, June 2-7, 2019, Volume
  1 (Long and Short Papers)}, pages 2924--2936. Association for Computational
  Linguistics.

\bibitem[{Clark et~al.(2020)Clark, Luong, Le, and Manning}]{clark2019electra}
Kevin Clark, Minh{-}Thang Luong, Quoc~V. Le, and Christopher~D. Manning. 2020.
\newblock \href {https://openreview.net/forum?id=r1xMH1BtvB} {{ELECTRA:}
  pre-training text encoders as discriminators rather than generators}.
\newblock In \emph{8th International Conference on Learning Representations,
  {ICLR} 2020, Addis Ababa, Ethiopia, April 26-30, 2020}. OpenReview.net.

\bibitem[{Dagan et~al.(2005)Dagan, Glickman, and Magnini}]{dagan2005pascal}
Ido Dagan, Oren Glickman, and Bernardo Magnini. 2005.
\newblock \href {https://doi.org/10.1007/11736790\_9} {The {PASCAL} recognising
  textual entailment challenge}.
\newblock In \emph{Machine Learning Challenges, Evaluating Predictive
  Uncertainty, Visual Object Classification and Recognizing Textual Entailment,
  First {PASCAL} Machine Learning Challenges Workshop, {MLCW} 2005,
  Southampton, UK, April 11-13, 2005, Revised Selected Papers}, volume 3944 of
  \emph{Lecture Notes in Computer Science}, pages 177--190. Springer.

\bibitem[{Davison et~al.(2019)Davison, Feldman, and
  Rush}]{davison2019commonsense}
Joe Davison, Joshua Feldman, and Alexander~M. Rush. 2019.
\newblock \href {https://doi.org/10.18653/v1/D19-1109} {Commonsense knowledge
  mining from pretrained models}.
\newblock In \emph{Proceedings of the 2019 Conference on Empirical Methods in
  Natural Language Processing and the 9th International Joint Conference on
  Natural Language Processing, {EMNLP-IJCNLP} 2019, Hong Kong, China, November
  3-7, 2019}, pages 1173--1178. Association for Computational Linguistics.

\bibitem[{Desai et~al.(2019)Desai, Zhan, and Aly}]{desai2019evaluating}
Shrey Desai, Hongyuan Zhan, and Ahmed Aly. 2019.
\newblock \href {https://doi.org/10.18653/v1/D19-6117} {Evaluating lottery
  tickets under distributional shifts}.
\newblock In \emph{Proceedings of the 2nd Workshop on Deep Learning Approaches
  for Low-Resource NLP, DeepLo@EMNLP-IJCNLP 2019, Hong Kong, China, November 3,
  2019}, pages 153--162. Association for Computational Linguistics.

\bibitem[{Devlin et~al.(2019)Devlin, Chang, Lee, and
  Toutanova}]{devlin2019bert}
Jacob Devlin, Ming{-}Wei Chang, Kenton Lee, and Kristina Toutanova. 2019.
\newblock \href {https://doi.org/10.18653/v1/n19-1423} {{BERT:} pre-training of
  deep bidirectional transformers for language understanding}.
\newblock In \emph{Proceedings of the 2019 Conference of the North American
  Chapter of the Association for Computational Linguistics: Human Language
  Technologies, {NAACL-HLT} 2019, Minneapolis, MN, USA, June 2-7, 2019, Volume
  1 (Long and Short Papers)}, pages 4171--4186. Association for Computational
  Linguistics.

\bibitem[{Ding et~al.(2021)Ding, Hu, Zhao, Chen, Liu, Zheng, and
  Sun}]{ding2021openprompt}
Ning Ding, Shengding Hu, Weilin Zhao, Yulin Chen, Zhiyuan Liu, Hai{-}Tao Zheng,
  and Maosong Sun. 2021.
\newblock \href {http://arxiv.org/abs/2111.01998} {Openprompt: An open-source
  framework for prompt-learning}.
\newblock \emph{CoRR}, abs/2111.01998.

\bibitem[{Ding et~al.(2022)Ding, Qin, Yang, Wei, Yang, Su, Hu, Chen, Chan,
  Chen, Yi, Zhao, Wang, Liu, Zheng, Chen, Liu, Tang, Li, and
  Sun}]{NNN2022DeltaTA}
Ning Ding, Yujia Qin, Guang Yang, Fu~Wei, Zonghan Yang, Yusheng Su, Shengding
  Hu, Yulin Chen, Chi-Min Chan, Weize Chen, Jing Yi, Weilin Zhao, Xiaozhi Wang,
  Zhiyuan Liu, Hai-Tao Zheng, Jianfei Chen, Yang Liu, Jie Tang, Juanzi Li, and
  Maosong Sun. 2022.
\newblock Delta tuning: A comprehensive study of parameter efficient methods
  for pre-trained language models.

\bibitem[{Frankle and Carbin(2019)}]{frankle2018lottery}
Jonathan Frankle and Michael Carbin. 2019.
\newblock \href {https://openreview.net/forum?id=rJl-b3RcF7} {The lottery
  ticket hypothesis: Finding sparse, trainable neural networks}.
\newblock In \emph{7th International Conference on Learning Representations,
  {ICLR} 2019, New Orleans, LA, USA, May 6-9, 2019}. OpenReview.net.

\bibitem[{Gong and Eldardiry(2021)}]{gong2021prompt}
Jiaying Gong and Hoda Eldardiry. 2021.
\newblock \href {http://arxiv.org/abs/2112.04539} {Prompt-based zero-shot
  relation classification with semantic knowledge augmentation}.
\newblock \emph{CoRR}, abs/2112.04539.

\bibitem[{Guo et~al.(2021)Guo, Rush, and Kim}]{guo2020parameter}
Demi Guo, Alexander~M. Rush, and Yoon Kim. 2021.
\newblock \href {https://doi.org/10.18653/v1/2021.acl-long.378}
  {Parameter-efficient transfer learning with diff pruning}.
\newblock In \emph{Proceedings of the 59th Annual Meeting of the Association
  for Computational Linguistics and the 11th International Joint Conference on
  Natural Language Processing, {ACL/IJCNLP} 2021, (Volume 1: Long Papers),
  Virtual Event, August 1-6, 2021}, pages 4884--4896. Association for
  Computational Linguistics.

\bibitem[{He et~al.(2022)He, Zheng, Tay, Gupta, Du, Aribandi, Zhao, Li, Chen,
  Metzler, Cheng, and Chi}]{he2022hyper}
Yun He, Huaixiu~Steven Zheng, Yi~Tay, Jai~Prakash Gupta, Yu~Du, Vamsi Aribandi,
  Zhe Zhao, YaGuang Li, Zhao Chen, Donald Metzler, Heng{-}Tze Cheng, and Ed~H.
  Chi. 2022.
\newblock Hyperprompt: Prompt-based task-conditioning of transformers.
\newblock \emph{CoRR}, abs/2203.00759.

\bibitem[{Houlsby et~al.(2019)Houlsby, Giurgiu, Jastrzebski, Morrone,
  de~Laroussilhe, Gesmundo, Attariyan, and Gelly}]{houlsby2019parameter}
Neil Houlsby, Andrei Giurgiu, Stanislaw Jastrzebski, Bruna Morrone, Quentin
  de~Laroussilhe, Andrea Gesmundo, Mona Attariyan, and Sylvain Gelly. 2019.
\newblock \href {http://proceedings.mlr.press/v97/houlsby19a.html}
  {Parameter-efficient transfer learning for {NLP}}.
\newblock In \emph{Proceedings of the 36th International Conference on Machine
  Learning, {ICML} 2019, 9-15 June 2019, Long Beach, California, {USA}},
  volume~97 of \emph{Proceedings of Machine Learning Research}, pages
  2790--2799. {PMLR}.

\bibitem[{Khashabi et~al.(2018)Khashabi, Chaturvedi, Roth, Upadhyay, and
  Roth}]{khashabi2018looking}
Daniel Khashabi, Snigdha Chaturvedi, Michael Roth, Shyam Upadhyay, and Dan
  Roth. 2018.
\newblock \href {https://doi.org/10.18653/v1/n18-1023} {Looking beyond the
  surface: {A} challenge set for reading comprehension over multiple
  sentences}.
\newblock In \emph{Proceedings of the 2018 Conference of the North American
  Chapter of the Association for Computational Linguistics: Human Language
  Technologies, {NAACL-HLT} 2018, New Orleans, Louisiana, USA, June 1-6, 2018,
  Volume 1 (Long Papers)}, pages 252--262. Association for Computational
  Linguistics.

\bibitem[{Khashabi et~al.(2020)Khashabi, Min, Khot, Sabharwal, Tafjord, Clark,
  and Hajishirzi}]{khashabi2020unifiedqa}
Daniel Khashabi, Sewon Min, Tushar Khot, Ashish Sabharwal, Oyvind Tafjord,
  Peter Clark, and Hannaneh Hajishirzi. 2020.
\newblock \href {https://doi.org/10.18653/v1/2020.findings-emnlp.171}
  {Unifiedqa: Crossing format boundaries with a single {QA} system}.
\newblock In \emph{Findings of the Association for Computational Linguistics:
  {EMNLP} 2020, Online Event, 16-20 November 2020}, volume {EMNLP} 2020 of
  \emph{Findings of {ACL}}, pages 1896--1907. Association for Computational
  Linguistics.

\bibitem[{Lester et~al.(2021)Lester, Al{-}Rfou, and Constant}]{lester2021power}
Brian Lester, Rami Al{-}Rfou, and Noah Constant. 2021.
\newblock \href {https://doi.org/10.18653/v1/2021.emnlp-main.243} {The power of
  scale for parameter-efficient prompt tuning}.
\newblock In \emph{Proceedings of the 2021 Conference on Empirical Methods in
  Natural Language Processing, {EMNLP} 2021, Virtual Event / Punta Cana,
  Dominican Republic, 7-11 November, 2021}, pages 3045--3059. Association for
  Computational Linguistics.

\bibitem[{Levesque et~al.(2012)Levesque, Davis, and
  Morgenstern}]{levesque2012winograd}
Hector~J. Levesque, Ernest Davis, and Leora Morgenstern. 2012.
\newblock \href {http://www.aaai.org/ocs/index.php/KR/KR12/paper/view/4492}
  {The winograd schema challenge}.
\newblock In \emph{Principles of Knowledge Representation and Reasoning:
  Proceedings of the Thirteenth International Conference, {KR} 2012, Rome,
  Italy, June 10-14, 2012}. {AAAI} Press.

\bibitem[{Lewis et~al.(2020)Lewis, Liu, Goyal, Ghazvininejad, Mohamed, Levy,
  Stoyanov, and Zettlemoyer}]{lewis2020bart}
Mike Lewis, Yinhan Liu, Naman Goyal, Marjan Ghazvininejad, Abdelrahman Mohamed,
  Omer Levy, Veselin Stoyanov, and Luke Zettlemoyer. 2020.
\newblock \href {https://doi.org/10.18653/v1/2020.acl-main.703} {{BART:}
  denoising sequence-to-sequence pre-training for natural language generation,
  translation, and comprehension}.
\newblock In \emph{Proceedings of the 58th Annual Meeting of the Association
  for Computational Linguistics, {ACL} 2020, Online, July 5-10, 2020}, pages
  7871--7880. Association for Computational Linguistics.

\bibitem[{Li and Liang(2021)}]{li2021prefix}
Xiang~Lisa Li and Percy Liang. 2021.
\newblock \href {https://doi.org/10.18653/v1/2021.acl-long.353} {Prefix-tuning:
  Optimizing continuous prompts for generation}.
\newblock In \emph{Proceedings of the 59th Annual Meeting of the Association
  for Computational Linguistics and the 11th International Joint Conference on
  Natural Language Processing, {ACL/IJCNLP} 2021, (Volume 1: Long Papers),
  Virtual Event, August 1-6, 2021}, pages 4582--4597. Association for
  Computational Linguistics.

\bibitem[{Liang et~al.(2021)Liang, Zuo, Chen, Jiang, Liu, He, Zhao, and
  Chen}]{liang2021super}
Chen Liang, Simiao Zuo, Minshuo Chen, Haoming Jiang, Xiaodong Liu, Pengcheng
  He, Tuo Zhao, and Weizhu Chen. 2021.
\newblock \href {https://doi.org/10.18653/v1/2021.acl-long.510} {Super tickets
  in pre-trained language models: From model compression to improving
  generalization}.
\newblock In \emph{Proceedings of the 59th Annual Meeting of the Association
  for Computational Linguistics and the 11th International Joint Conference on
  Natural Language Processing, {ACL/IJCNLP} 2021, (Volume 1: Long Papers),
  Virtual Event, August 1-6, 2021}, pages 6524--6538. Association for
  Computational Linguistics.

\bibitem[{Liu et~al.(2021{\natexlab{a}})Liu, Yuan, Fu, Jiang, Hayashi, and
  Neubig}]{liu2021pre}
Pengfei Liu, Weizhe Yuan, Jinlan Fu, Zhengbao Jiang, Hiroaki Hayashi, and
  Graham Neubig. 2021{\natexlab{a}}.
\newblock \href {http://arxiv.org/abs/2107.13586} {Pre-train, prompt, and
  predict: {A} systematic survey of prompting methods in natural language
  processing}.
\newblock \emph{CoRR}, abs/2107.13586.

\bibitem[{Liu et~al.(2021{\natexlab{b}})Liu, Ji, Fu, Du, Yang, and
  Tang}]{liu2021p}
Xiao Liu, Kaixuan Ji, Yicheng Fu, Zhengxiao Du, Zhilin Yang, and Jie Tang.
  2021{\natexlab{b}}.
\newblock \href {http://arxiv.org/abs/2110.07602} {P-tuning v2: Prompt tuning
  can be comparable to fine-tuning universally across scales and tasks}.
\newblock \emph{CoRR}, abs/2110.07602.

\bibitem[{Liu et~al.(2021{\natexlab{c}})Liu, Zheng, Du, Ding, Qian, Yang, and
  Tang}]{liu2021gpt}
Xiao Liu, Yanan Zheng, Zhengxiao Du, Ming Ding, Yujie Qian, Zhilin Yang, and
  Jie Tang. 2021{\natexlab{c}}.
\newblock \href {http://arxiv.org/abs/2103.10385} {{GPT} understands, too}.
\newblock \emph{CoRR}, abs/2103.10385.

\bibitem[{Liu et~al.(2019)Liu, Ott, Goyal, Du, Joshi, Chen, Levy, Lewis,
  Zettlemoyer, and Stoyanov}]{liu2019roberta}
Yinhan Liu, Myle Ott, Naman Goyal, Jingfei Du, Mandar Joshi, Danqi Chen, Omer
  Levy, Mike Lewis, Luke Zettlemoyer, and Veselin Stoyanov. 2019.
\newblock \href {http://arxiv.org/abs/1907.11692} {Roberta: {A} robustly
  optimized {BERT} pretraining approach}.
\newblock \emph{CoRR}, abs/1907.11692.

\bibitem[{Michel et~al.(2019)Michel, Levy, and Neubig}]{michel2019sixteen}
Paul Michel, Omer Levy, and Graham Neubig. 2019.
\newblock \href
  {https://proceedings.neurips.cc/paper/2019/hash/2c601ad9d2ff9bc8b282670cdd54f69f-Abstract.html}
  {Are sixteen heads really better than one?}
\newblock In \emph{Advances in Neural Information Processing Systems 32: Annual
  Conference on Neural Information Processing Systems 2019, NeurIPS 2019,
  December 8-14, 2019, Vancouver, BC, Canada}, pages 14014--14024.

\bibitem[{Morcos et~al.(2019)Morcos, Yu, Paganini, and Tian}]{morcos2019one}
Ari~S. Morcos, Haonan Yu, Michela Paganini, and Yuandong Tian. 2019.
\newblock \href
  {https://proceedings.neurips.cc/paper/2019/hash/a4613e8d72a61b3b69b32d040f89ad81-Abstract.html}
  {One ticket to win them all: generalizing lottery ticket initializations
  across datasets and optimizers}.
\newblock In \emph{Advances in Neural Information Processing Systems 32: Annual
  Conference on Neural Information Processing Systems 2019, NeurIPS 2019,
  December 8-14, 2019, Vancouver, BC, Canada}, pages 4933--4943.

\bibitem[{Pilehvar and Camacho{-}Collados(2019)}]{pilehvar2019wic}
Mohammad~Taher Pilehvar and Jos{\'{e}} Camacho{-}Collados. 2019.
\newblock \href {https://doi.org/10.18653/v1/n19-1128} {Wic: the
  word-in-context dataset for evaluating context-sensitive meaning
  representations}.
\newblock In \emph{Proceedings of the 2019 Conference of the North American
  Chapter of the Association for Computational Linguistics: Human Language
  Technologies, {NAACL-HLT} 2019, Minneapolis, MN, USA, June 2-7, 2019, Volume
  1 (Long and Short Papers)}, pages 1267--1273. Association for Computational
  Linguistics.

\bibitem[{Qin and Eisner(2021)}]{qin2021learning}
Guanghui Qin and Jason Eisner. 2021.
\newblock \href {https://doi.org/10.18653/v1/2021.naacl-main.410} {Learning how
  to ask: Querying lms with mixtures of soft prompts}.
\newblock In \emph{Proceedings of the 2021 Conference of the North American
  Chapter of the Association for Computational Linguistics: Human Language
  Technologies, {NAACL-HLT} 2021, Online, June 6-11, 2021}, pages 5203--5212.
  Association for Computational Linguistics.

\bibitem[{Radford et~al.(2019)Radford, Wu, Child, Luan, Amodei, Sutskever
  et~al.}]{radford2019language}
Alec Radford, Jeffrey Wu, Rewon Child, David Luan, Dario Amodei, Ilya
  Sutskever, et~al. 2019.
\newblock Language models are unsupervised multitask learners.
\newblock \emph{OpenAI blog}, 1(8):9.

\bibitem[{Raffel et~al.(2020)Raffel, Shazeer, Roberts, Lee, Narang, Matena,
  Zhou, Li, and Liu}]{raffel2020exploring}
Colin Raffel, Noam Shazeer, Adam Roberts, Katherine Lee, Sharan Narang, Michael
  Matena, Yanqi Zhou, Wei Li, and Peter~J. Liu. 2020.
\newblock \href {http://jmlr.org/papers/v21/20-074.html} {Exploring the limits
  of transfer learning with a unified text-to-text transformer}.
\newblock \emph{J. Mach. Learn. Res.}, 21:140:1--140:67.

\bibitem[{Renda et~al.(2020)Renda, Frankle, and Carbin}]{renda2019comparing}
Alex Renda, Jonathan Frankle, and Michael Carbin. 2020.
\newblock \href {https://openreview.net/forum?id=S1gSj0NKvB} {Comparing
  rewinding and fine-tuning in neural network pruning}.
\newblock In \emph{8th International Conference on Learning Representations,
  {ICLR} 2020, Addis Ababa, Ethiopia, April 26-30, 2020}. OpenReview.net.

\bibitem[{Roemmele et~al.(2011)Roemmele, Bejan, and
  Gordon}]{roemmele2011choice}
Melissa Roemmele, Cosmin~Adrian Bejan, and Andrew~S. Gordon. 2011.
\newblock \href {http://www.aaai.org/ocs/index.php/SSS/SSS11/paper/view/2418}
  {Choice of plausible alternatives: An evaluation of commonsense causal
  reasoning}.
\newblock In \emph{Logical Formalizations of Commonsense Reasoning, Papers from
  the 2011 {AAAI} Spring Symposium, Technical Report SS-11-06, Stanford,
  California, USA, March 21-23, 2011}. {AAAI}.

\bibitem[{Schick and Sch{\"{u}}tze(2021)}]{schick2021s}
Timo Schick and Hinrich Sch{\"{u}}tze. 2021.
\newblock \href {https://doi.org/10.18653/v1/2021.naacl-main.185} {It's not
  just size that matters: Small language models are also few-shot learners}.
\newblock In \emph{Proceedings of the 2021 Conference of the North American
  Chapter of the Association for Computational Linguistics: Human Language
  Technologies, {NAACL-HLT} 2021, Online, June 6-11, 2021}, pages 2339--2352.
  Association for Computational Linguistics.

\bibitem[{Shazeer and Stern(2018)}]{shazeer2018adafactor}
Noam Shazeer and Mitchell Stern. 2018.
\newblock \href {http://proceedings.mlr.press/v80/shazeer18a.html} {Adafactor:
  Adaptive learning rates with sublinear memory cost}.
\newblock In \emph{Proceedings of the 35th International Conference on Machine
  Learning, {ICML} 2018, Stockholmsm{\"{a}}ssan, Stockholm, Sweden, July 10-15,
  2018}, volume~80 of \emph{Proceedings of Machine Learning Research}, pages
  4603--4611. {PMLR}.

\bibitem[{Simonyan et~al.(2014)Simonyan, Vedaldi, and
  Zisserman}]{simonyan2014deep}
Karen Simonyan, Andrea Vedaldi, and Andrew Zisserman. 2014.
\newblock \href {http://arxiv.org/abs/1312.6034} {Deep inside convolutional
  networks: Visualising image classification models and saliency maps}.
\newblock In \emph{2nd International Conference on Learning Representations,
  {ICLR} 2014, Banff, AB, Canada, April 14-16, 2014, Workshop Track
  Proceedings}.

\bibitem[{Vu et~al.(2021)Vu, Lester, Constant, Al{-}Rfou, and Cer}]{vu2021spot}
Tu~Vu, Brian Lester, Noah Constant, Rami Al{-}Rfou, and Daniel Cer. 2021.
\newblock \href {http://arxiv.org/abs/2110.07904} {Spot: Better frozen model
  adaptation through soft prompt transfer}.
\newblock \emph{CoRR}, abs/2110.07904.

\bibitem[{Wang et~al.(2019)Wang, Pruksachatkun, Nangia, Singh, Michael, Hill,
  Levy, and Bowman}]{wang2019superglue}
Alex Wang, Yada Pruksachatkun, Nikita Nangia, Amanpreet Singh, Julian Michael,
  Felix Hill, Omer Levy, and Samuel~R. Bowman. 2019.
\newblock \href
  {https://proceedings.neurips.cc/paper/2019/hash/4496bf24afe7fab6f046bf4923da8de6-Abstract.html}
  {Superglue: {A} stickier benchmark for general-purpose language understanding
  systems}.
\newblock In \emph{Advances in Neural Information Processing Systems 32: Annual
  Conference on Neural Information Processing Systems 2019, NeurIPS 2019,
  December 8-14, 2019, Vancouver, BC, Canada}, pages 3261--3275.

\bibitem[{Wang et~al.(2021)Wang, Wang, Qiu, Huang, and
  Gao}]{wang2021transprompt}
Chengyu Wang, Jianing Wang, Minghui Qiu, Jun Huang, and Ming Gao. 2021.
\newblock \href {https://doi.org/10.18653/v1/2021.emnlp-main.221} {Transprompt:
  Towards an automatic transferable prompting framework for few-shot text
  classification}.
\newblock In \emph{Proceedings of the 2021 Conference on Empirical Methods in
  Natural Language Processing, {EMNLP} 2021, Virtual Event / Punta Cana,
  Dominican Republic, 7-11 November, 2021}, pages 2792--2802. Association for
  Computational Linguistics.

\bibitem[{Yang et~al.(2019)Yang, Dai, Yang, Carbonell, Salakhutdinov, and
  Le}]{yang2019xlnet}
Zhilin Yang, Zihang Dai, Yiming Yang, Jaime~G. Carbonell, Ruslan Salakhutdinov,
  and Quoc~V. Le. 2019.
\newblock \href
  {https://proceedings.neurips.cc/paper/2019/hash/dc6a7e655d7e5840e66733e9ee67cc69-Abstract.html}
  {Xlnet: Generalized autoregressive pretraining for language understanding}.
\newblock In \emph{Advances in Neural Information Processing Systems 32: Annual
  Conference on Neural Information Processing Systems 2019, NeurIPS 2019,
  December 8-14, 2019, Vancouver, BC, Canada}, pages 5754--5764.

\bibitem[{Yu et~al.(2020)Yu, Edunov, Tian, and Morcos}]{yu2019playing}
Haonan Yu, Sergey Edunov, Yuandong Tian, and Ari~S. Morcos. 2020.
\newblock \href {https://openreview.net/forum?id=S1xnXRVFwH} {Playing the
  lottery with rewards and multiple languages: lottery tickets in {RL} and
  {NLP}}.
\newblock In \emph{8th International Conference on Learning Representations,
  {ICLR} 2020, Addis Ababa, Ethiopia, April 26-30, 2020}. OpenReview.net.

\bibitem[{Zhang et~al.(2018)Zhang, Liu, Liu, Gao, Duh, and
  Durme}]{zhang2018record}
Sheng Zhang, Xiaodong Liu, Jingjing Liu, Jianfeng Gao, Kevin Duh, and
  Benjamin~Van Durme. 2018.
\newblock \href {http://arxiv.org/abs/1810.12885} {Record: Bridging the gap
  between human and machine commonsense reading comprehension}.
\newblock \emph{CoRR}, abs/1810.12885.

\bibitem[{Zhong et~al.(2021)Zhong, Friedman, and Chen}]{zhong2021factual}
Zexuan Zhong, Dan Friedman, and Danqi Chen. 2021.
\newblock \href {https://doi.org/10.18653/v1/2021.naacl-main.398} {Factual
  probing is {[MASK]:} learning vs. learning to recall}.
\newblock In \emph{Proceedings of the 2021 Conference of the North American
  Chapter of the Association for Computational Linguistics: Human Language
  Technologies, {NAACL-HLT} 2021, Online, June 6-11, 2021}, pages 5017--5033.
  Association for Computational Linguistics.

\bibitem[{Zhou et~al.(2020)Zhou, Zhou, Zhao, Wang, and Wen}]{zhou2020towards}
Kun Zhou, Yuanhang Zhou, Wayne~Xin Zhao, Xiaoke Wang, and Ji{-}Rong Wen. 2020.
\newblock \href {https://doi.org/10.18653/v1/2020.coling-main.365} {Towards
  topic-guided conversational recommender system}.
\newblock In \emph{Proceedings of the 28th International Conference on
  Computational Linguistics, {COLING} 2020, Barcelona, Spain (Online), December
  8-13, 2020}, pages 4128--4139. International Committee on Computational
  Linguistics.

\end{thebibliography}
\bibliographystyle{acl_natbib}
\clearpage

\appendix

\section{More Results of P-TuningV2}
We observe that the performance of Prompt-Tuning and P-Tuning are comparable at the same model scale. Moreover, P-TuningV$2$ outperforms Prompt-Tuning and P-Tuning on CB, RTE, and Boolq. However, \textsc{XPrompt} achieves more predominant performances than P-TuningV$2$ at similar model scales, demonstrating its effectiveness.
\begin{table}[ht]
\centering
\resizebox{0.42\textwidth}{!}{
\begin{tabular}{cc|ccc}
\toprule
\multicolumn{2}{c|}{\textbf{Model}}                                                                       & \textbf{CB} & \textbf{RTE}   & \textbf{Boolq} \\ \hline
\multicolumn{1}{c|}{\multirow{3}{*}{\begin{tabular}[c]{@{}c@{}}GLM-XL\\ 2B\end{tabular}}}   & Fine-Tuning$^\dag$ & 96.40       & 90.30 & 88.30          \\ 

\multicolumn{1}{c|}{}                                                                       & P-Tuning$^\dag$    & 76.40       & 85.60 & 79.70          \\
\multicolumn{1}{c|}{}                                                                       & P-TuningV2$^\dag$  & \textbf{96.40}       & 90.30 & 87.00          \\ \hline

\multicolumn{1}{c|}{\begin{tabular}[c]{@{}c@{}}T5-XL 3B\end{tabular}}                 & \textbf{\textsc{XPrompt}}     & 92.85       &\textbf{92.79} & \textbf{89.00}        \\ \hline

\multicolumn{1}{c|}{\multirow{3}{*}{\begin{tabular}[c]{@{}c@{}}GLM-XXL\\ 10B\end{tabular}}} & Fine-Tuning$^\dag$ & 98.70       & 93.10 & 88.70          \\
\multicolumn{1}{c|}{}                                                                       & P-Tuning$^\dag$    & 98.20       & 89.90 & 88.80          \\
\multicolumn{1}{c|}{}                                                                       & P-TuningV2$^\dag$  & 96.40       & 93.10 & 88.80          \\ \hline

\multicolumn{1}{c|}{\begin{tabular}[c]{@{}c@{}}T5-XXL 11B\end{tabular}}               & \textbf{\textsc{XPrompt}}     &\textbf{100.00}       &\textbf{94.94} &\textbf{90.87}          \\  \bottomrule
\end{tabular}}
\caption{The results on three SuperGLUE tasks for different models and similar model scales. The better results are in bold. Methods with `$^\dag$' indicate results reported in ~\citet{liu2021p}. \textsc{XPrompt} surpasses P-tuningV2 on models with similar scales.}
 \label{PtuningV2results}
\end{table}

\section{Token and Piece Importance Score Distribution}
\begin{figure}[ht]
  \centering
  \resizebox{0.38\textwidth}{!}{
  \includegraphics[width=\linewidth]{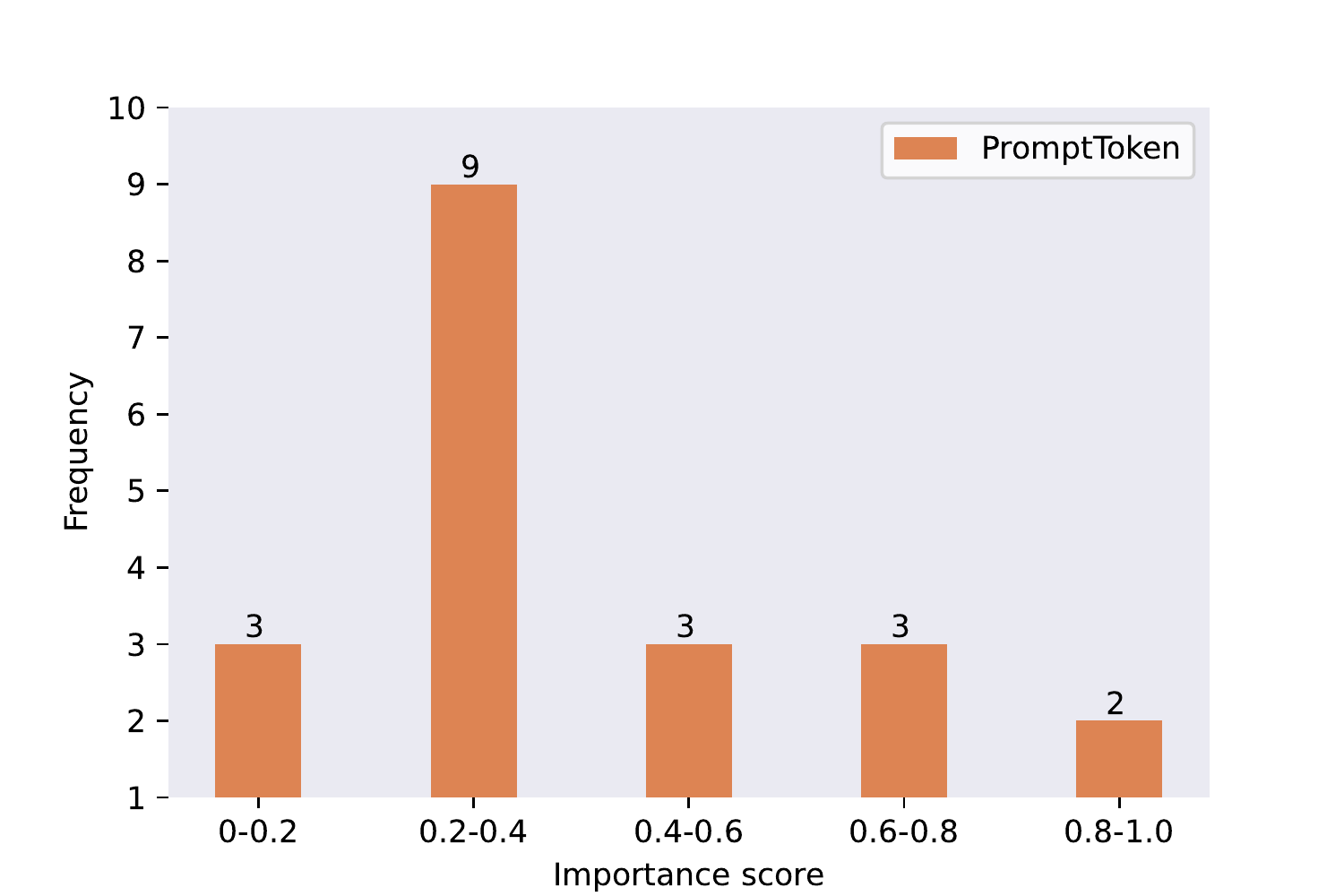}}
  \caption{The distribution of prompt tokens' importance scores on WSC task.}
  \label{PromptTokensImportanceScore}  
\end{figure}

\begin{figure}[ht]
  \centering
  \resizebox{0.40\textwidth}{!}{
  \includegraphics[width=\linewidth]{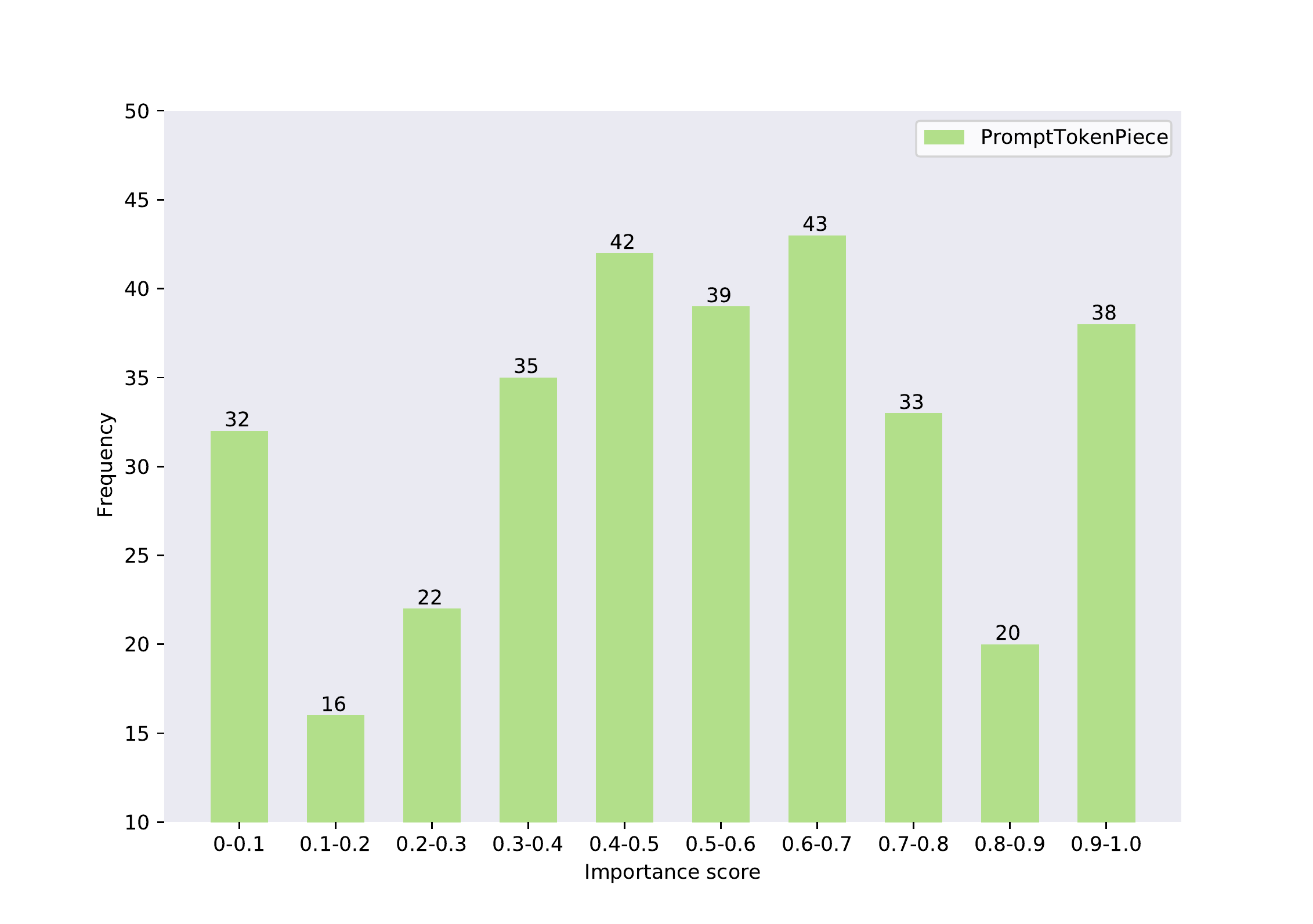}}
  \caption{The distribution of prompt token pieces' importance scores on WSC task. }
  \label{PromptTokenPiecesImportanceScore} 
\end{figure}

Figure~\ref{PromptTokensImportanceScore} and  Figure~\ref{PromptTokenPiecesImportanceScore} show the distribution of prompt tokens' and prompt token pieces' importance scores on the WSC task. It is clear that most prompt tokens have a low importance score, and only a few prompt tokens have a large importance score. These results further demonstrate our hypothesis that the existence of negative prompts, and their stability.

\section{\textsc{XPrompt} Transfer}
\label{section:XPrompt_Transfer}
\begin{figure}[ht]
  \centering
  \includegraphics[width=\linewidth]{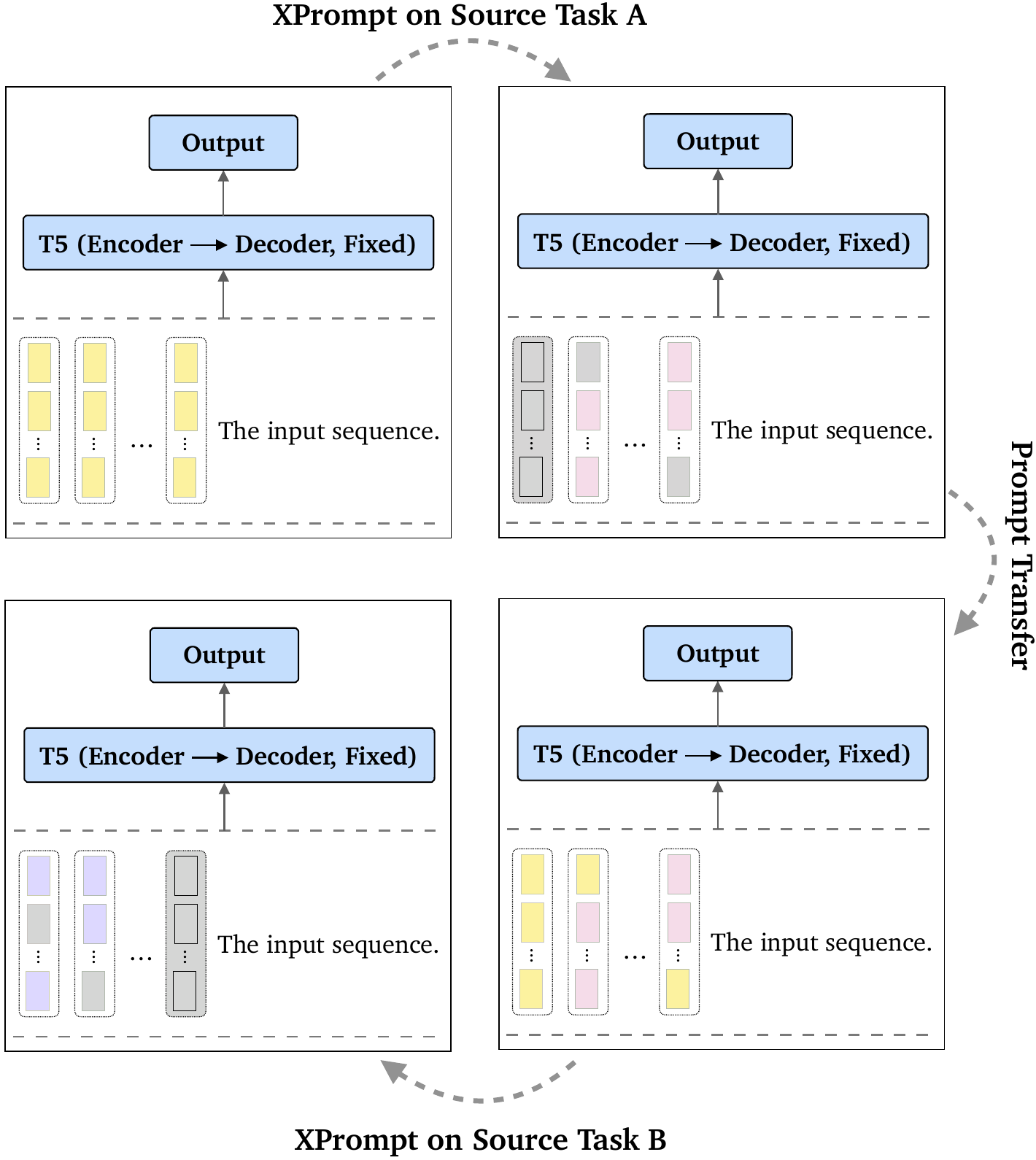}
  \caption{The illustration of \textsc{XPrompt} Transfer approach. \textsc{XPrompt} Transfer first trains the prompts through \textsc{XPrompt} on the source task A and then uses the resulting prompts to initialize the prompts of the target task B, followed by the XPrompt training on the target task B.}
  \label{XPromptTransfer}  
\end{figure}

As shown in Figure~\ref{XPromptTransfer}, given a source task and a target task, \textsc{XPrompt} Transfer first trains the prompts through our \textsc{XPrompt} on the source task and then uses the resulting prompts to initialize the prompts of the target task, followed by the XPrompt training on the target task. Different from \textsc{SPoT}, we do not use the trained prompts to initialize the prompts for the target task, and our approach can provide more cross tasks information to the prompts. The results of  different prompt initialization methods are shown in Table~\ref{initializationResults}, and the results of \textsc{XPrompt} Transfer are shown in Table~\ref{PromptTransfer_results}.

\clearpage

\section{Importance Scores Visualization}
\label{section:importance_scores_masked_visua}
Figure~\ref{visual_importance_scores_piece_original} and Figure~\ref{visual_importance_scores_piece_masked} show the original and pruned gradient saliency maps of the importance scores on WSC task. The gray cells in Figure~\ref{visual_importance_scores_piece_masked} indicate that the prompt tokens and pieces are pruned due to low importance scores, and the remaining ones are the winning tickets.

\begin{figure}[ht]
\centering
\resizebox{0.44\textwidth}{!}{
\begin{tabular}{cc}
\toprule  
\textbf{Tokens} & \textbf{Soft Prompt Token Pieces}     \\  
\hline
\colorbox{red!63.552068}{\strut T0} 
& 
\colorbox{red!89.36}{\strut P0}
\colorbox{red!61.50}{\strut P1} 
\colorbox{red!100.0}{\strut P2} 
\colorbox{red!45.59}{\strut P3} 
\colorbox{red!37.12}{\strut P4} 
\colorbox{red!37.42}{\strut P5} 
\colorbox{red!32.45}{\strut P6}
\colorbox{red!73.51}{\strut P7} 
\colorbox{red!0.0}{\strut P8} 
\colorbox{red!46.29}{\strut P9} 
\colorbox{red!31.89}{\strut P10}
\colorbox{red!37.84}{\strut P11} 
\colorbox{red!55.67}{\strut P12} 
\colorbox{red!70.84}{\strut P13} 
\colorbox{red!67.18}{\strut P14} 
\colorbox{red!35.87}{\strut P15} 

\\ 
\colorbox{red!0.0}{\strut T1} 
& 
\colorbox{red!64.82}{\strut P0}
\colorbox{red!64.81}{\strut P1} 
\colorbox{red!52.61}{\strut P2} 
\colorbox{red!31.98}{\strut P3} 
\colorbox{red!85.44}{\strut P4} 
\colorbox{red!66.05}{\strut P5} 
\colorbox{red!60.16}{\strut P6}
\colorbox{red!95.54}{\strut P7} 
\colorbox{red!73.17}{\strut P8} 
\colorbox{red!49.39}{\strut P9} 
\colorbox{red!06.75}{\strut P10}
\colorbox{red!51.10}{\strut P11} 
\colorbox{red!100.0}{\strut P12} 
\colorbox{red!6.54}{\strut P13} 
\colorbox{red!28.63}{\strut P14} 
\colorbox{red!0.0}{\strut P15} 
\\
\colorbox{red!61.578697}{\strut T2} 
& 
\colorbox{red!51.10}{\strut P0}
\colorbox{red!31.29}{\strut P1} 
\colorbox{red!42.23}{\strut P2} 
\colorbox{red!42.10}{\strut P3} 
\colorbox{red!1.53}{\strut P4} 
\colorbox{red!15.27}{\strut P5} 
\colorbox{red!87.46}{\strut P6}
\colorbox{red!9.83}{\strut P7} 
\colorbox{red!57.96}{\strut P8} 
\colorbox{red!4.45}{\strut P9} 
\colorbox{red!71.28085821377794}{\strut P10}
\colorbox{red!99.99999999999998}{\strut P11} 
\colorbox{red!99.84490112446684}{\strut P12} 
\colorbox{red!0.0}{\strut P13} 
\colorbox{red!40.33863254491405}{\strut P14} 
\colorbox{red!39.20124079100426}{\strut P15} 
\\
\colorbox{red!32.810271}{\strut T3}
& 
\colorbox{red!85.41362174802689}{\strut P0}
\colorbox{red!60.07989866510767}{\strut P1} 
\colorbox{red!46.51661307609861}{\strut P2} 
\colorbox{red!77.91094221962389}{\strut P3} 
\colorbox{red!99.34716944363247}{\strut P4} 
\colorbox{red!54.42852966968723}{\strut P5} 
\colorbox{red!52.54798791776284}{\strut P6}
\colorbox{red!100.0}{\strut P7} 
\colorbox{red!51.57361395303518}{\strut P8} 
\colorbox{red!53.7756991133197}{\strut P9} 
\colorbox{red!31.053298255870604}{\strut P10}
\colorbox{red!46.565331774334995}{\strut P11} 
\colorbox{red!60.91786027477346}{\strut P12} 
\colorbox{red!45.951476176556566}{\strut P13} 
\colorbox{red!68.7810581701257}{\strut P14} 
\colorbox{red!0.0}{\strut P15} 
\\
\colorbox{red!32.477413}{\strut T4}
& 
\colorbox{red!42.42481086030778}{\strut P0}
\colorbox{red!62.31875038894767}{\strut P1} 
\colorbox{red!33.05192784687464}{\strut P2} 
\colorbox{red!100.0}{\strut P3} 
\colorbox{red!19.75765226745375}{\strut P4} 
\colorbox{red!36.928069130446384}{\strut P5} 
\colorbox{red!37.843674155205675}{\strut P6}
\colorbox{red!36.11939581981277}{\strut P7} 
\colorbox{red!12.448992292168595}{\strut P8} 
\colorbox{red!29.61007441124437}{\strut P9} 
\colorbox{red!47.99303005787542}{\strut P10}
\colorbox{red!22.747437390539016}{\strut P11} 
\colorbox{red!35.684944391596957}{\strut P12} 
\colorbox{red!43.18216975009557}{\strut P13} 
\colorbox{red!0.0}{\strut P14} 
\colorbox{red!12.87572344265356}{\strut P15} 
\\
\colorbox{red!50.237756}{\strut T5}
& 
\colorbox{red!19.92481203007519}{\strut P0}
\colorbox{red!91.04856736435685}{\strut P1} 
\colorbox{red!59.62202804308068}{\strut P2} 
\colorbox{red!60.07925218451534}{\strut P3} 
\colorbox{red!0.0}{\strut P4} 
\colorbox{red!31.223328591749644}{\strut P5} 
\colorbox{red!44.65555781345255}{\strut P6}
\colorbox{red!100.0}{\strut P7} 
\colorbox{red!17.943507417191626}{\strut P8} 
\colorbox{red!51.30054866896973}{\strut P9} 
\colorbox{red!38.91485470432839}{\strut P10}
\colorbox{red!40.93680146311725}{\strut P11} 
\colorbox{red!37.492379597642755}{\strut P12} 
\colorbox{red!51.69680959154643}{\strut P13} 
\colorbox{red!30.674659622028044}{\strut P14} 
\colorbox{red!21.255842308473888}{\strut P15} 
\\
\colorbox{red!38.456966}{\strut T6}
& 
\colorbox{red!71.39838408949657}{\strut P0}
\colorbox{red!00.6463642013672999}{\strut P1} 
\colorbox{red!89.7327532628962}{\strut P2} 
\colorbox{red!58.8688626476072}{\strut P3} 
\colorbox{red!30.47855811062771}{\strut P4} 
\colorbox{red!44.474829086389056}{\strut P5} 
\colorbox{red!29.72032318210068}{\strut P6}
\colorbox{red!73.93412057178371}{\strut P7} 
\colorbox{red!61.08141702921068}{\strut P8} 
\colorbox{red!69.57116221255437}{\strut P9} 
\colorbox{red!0.0}{\strut P10}
\colorbox{red!13.026724673710366}{\strut P11} 
\colorbox{red!100.0}{\strut P12} 
\colorbox{red!93.53635798632691}{\strut P13} 
\colorbox{red!77.83716594157861}{\strut P14} 
\colorbox{red!43.244251087632063}{\strut P15} 
\\
\colorbox{red!77.104137}{\strut T7}
& 
\colorbox{red!95.56994160954556}{\strut P0}
\colorbox{red!35.05965981213506}{\strut P1} 
\colorbox{red!38.499619192688495}{\strut P2} 
\colorbox{red!66.33663366336633}{\strut P3} 
\colorbox{red!66.55242447321654}{\strut P4} 
\colorbox{red!09.634424980959633}{\strut P5} 
\colorbox{red!96.4077176948464}{\strut P6}
\colorbox{red!0.0}{\strut P7} 
\colorbox{red!86.17669459253616}{\strut P8} 
\colorbox{red!19.39578573241939}{\strut P9} 
\colorbox{red!56.61335364305661}{\strut P10}
\colorbox{red!55.21706016755521}{\strut P11} 
\colorbox{red!09.063214013709064}{\strut P12} 
\colorbox{red!74.09240924092408}{\strut P13} 
\colorbox{red!56.75298299060675}{\strut P14} 
\colorbox{red!99.99999999999999}{\strut P15} 
\\
\colorbox{red!26.842606}{\strut T8}
& 
\colorbox{red!41.21247113163972}{\strut P0}
\colorbox{red!38.706697459584294}{\strut P1} 
\colorbox{red!100.0}{\strut P2} 
\colorbox{red!50.99307159353348}{\strut P3} 
\colorbox{red!88.51039260969976}{\strut P4} 
\colorbox{red!35.36951501154734}{\strut P5} 
\colorbox{red!70.11547344110854}{\strut P6}
\colorbox{red!66.10854503464203}{\strut P7} 
\colorbox{red!03.23325635103926}{\strut P8} 
\colorbox{red!0.0}{\strut P9} 
\colorbox{red!76.25866050808314}{\strut P10}
\colorbox{red!79.04157043879907}{\strut P11} 
\colorbox{red!66.48960739030023}{\strut P12} 
\colorbox{red!84.67667436489607}{\strut P13} 
\colorbox{red!83.70669745958429}{\strut P14} 
\colorbox{red!47.66743648960739}{\strut P15} 
\\
\colorbox{red!59.938184}{\strut T9}
& 
\colorbox{red!78.78837228238743}{\strut P0}
\colorbox{red!65.79268789186549}{\strut P1} 
\colorbox{red!70.1490106668838}{\strut P2} 
\colorbox{red!66.83494829411286}{\strut P3} 
\colorbox{red!68.93575441739273}{\strut P4} 
\colorbox{red!44.93119452813289}{\strut P5} 
\colorbox{red!0.0}{\strut P6}
\colorbox{red!62.29134435306571}{\strut P7} 
\colorbox{red!71.48440680726326}{\strut P8} 
\colorbox{red!71.34598159758978}{\strut P9} 
\colorbox{red!100.0}{\strut P10}
\colorbox{red!69.64416578454523}{\strut P11} 
\colorbox{red!64.62014493933719}{\strut P12} 
\colorbox{red!54.1486849605081}{\strut P13} 
\colorbox{red!81.31259669408029}{\strut P14} 
\colorbox{red!65.2715576907418}{\strut P15} 
\\
\colorbox{red!25.32097}{\strut T10}
& 
\colorbox{red!67.70675552862768}{\strut P0}
\colorbox{red!48.389376956477836}{\strut P1} 
\colorbox{red!28.264162375037866}{\strut P2} 
\colorbox{red!57.56841361203675}{\strut P3} 
\colorbox{red!09.582954660203977}{\strut P4} 
\colorbox{red!30.980510956275875}{\strut P5} 
\colorbox{red!63.59688983136422}{\strut P6}
\colorbox{red!35.36302130667474}{\strut P7} 
\colorbox{red!23.326264768252042}{\strut P8} 
\colorbox{red!45.79420377663334}{\strut P9} 
\colorbox{red!18.852872866808038}{\strut P10}
\colorbox{red!0.0}{\strut P11} 
\colorbox{red!18.620620014137126}{\strut P12} 
\colorbox{red!99.99999999999999}{\strut P13} 
\colorbox{red!43.31010804806624}{\strut P14} 
\colorbox{red!21.407654246188024}{\strut P15} 
\\
\colorbox{red!19.626724}{\strut T11} 
& 
\colorbox{red!100.0}{\strut P0}
\colorbox{red!67.16938545193393}{\strut P1} 
\colorbox{red!39.468554427003183}{\strut P2} 
\colorbox{red!53.88324612701344}{\strut P3} 
\colorbox{red!54.32440751000308}{\strut P4} 
\colorbox{red!73.15071303990972}{\strut P5} 
\colorbox{red!94.59320816661538}{\strut P6}
\colorbox{red!91.925720734585}{\strut P7} 
\colorbox{red!27.177593105570946}{\strut P8} 
\colorbox{red!79.09100235970042}{\strut P9} 
\colorbox{red!80.79409048938135}{\strut P10}
\colorbox{red!50.12824458807839}{\strut P11} 
\colorbox{red!52.65209808146096}{\strut P12} 
\colorbox{red!53.44208474402381}{\strut P13} 
\colorbox{red!50.99004821996512}{\strut P14} 
\colorbox{red!0.0}{\strut P15} 
\\
\colorbox{red!30.432715}{\strut T12}
& 
\colorbox{red!36.57321480902608}{\strut P0}
\colorbox{red!56.14926247924196}{\strut P1} 
\colorbox{red!44.75920679886686}{\strut P2} 
\colorbox{red!49.223405294519873}{\strut P3} 
\colorbox{red!47.504151606916084}{\strut P4} 
\colorbox{red!52.74982905147992}{\strut P5} 
\colorbox{red!19.449057341017872}{\strut P6}
\colorbox{red!64.89205822018169}{\strut P7} 
\colorbox{red!72.28680277425026}{\strut P8} 
\colorbox{red!82.19204845169483}{\strut P9} 
\colorbox{red!57.08703721793493}{\strut P10}
\colorbox{red!81.46918042395233}{\strut P11} 
\colorbox{red!99.99999999999999}{\strut P12} 
\colorbox{red!76.99521344143792}{\strut P13} 
\colorbox{red!41.92634560906515}{\strut P14} 
\colorbox{red!0.0}{\strut P15} 
\\
\colorbox{red!57.310984}{\strut T13} 
& 
\colorbox{red!65.45201115093588}{\strut P0}
\colorbox{red!71.46555157307847}{\strut P1} 
\colorbox{red!96.30625248904819}{\strut P2} 
\colorbox{red!72.55077658303466}{\strut P3} 
\colorbox{red!80.56551174830746}{\strut P4} 
\colorbox{red!43.65790521704501}{\strut P5} 
\colorbox{red!70.84826762246118}{\strut P6}
\colorbox{red!83.92074870569495}{\strut P7} 
\colorbox{red!100.0}{\strut P8} 
\colorbox{red!66.43767423337317}{\strut P9} 
\colorbox{red!54.56989247311829}{\strut P10}
\colorbox{red!81.2226204699323}{\strut P11} 
\colorbox{red!98.40700915969733}{\strut P12} 
\colorbox{red!0.0}{\strut P13} 
\colorbox{red!58.28355236957389}{\strut P14} 
\colorbox{red!43.10035842293908}{\strut P15} 
\\
\colorbox{red!100.0}{\strut T14}
& 
\colorbox{red!21.754746023601845}{\strut P0}
\colorbox{red!24.5638789122627}{\strut P1} 
\colorbox{red!91.91893278604413}{\strut P2} 
\colorbox{red!69.79220112878399}{\strut P3} 
\colorbox{red!54.82298614674193}{\strut P4} 
\colorbox{red!72.90918419702412}{\strut P5} 
\colorbox{red!23.178553104155974}{\strut P6}
\colorbox{red!98.51205746536686}{\strut P7} 
\colorbox{red!0.0}{\strut P8} 
\colorbox{red!55.50282195997948}{\strut P9} 
\colorbox{red!69.97178040020524}{\strut P10}
\colorbox{red!45.08722421754746}{\strut P11} 
\colorbox{red!69.75371985633658}{\strut P12} 
\colorbox{red!100.0}{\strut P13} 
\colorbox{red!56.18265777321704}{\strut P14} 
\colorbox{red!03.501795792714213}{\strut P15} 
\\
\colorbox{red!14.110794}{\strut T15} 
& 
\colorbox{red!93.9804182279705}{\strut P0}
\colorbox{red!100.0}{\strut P1} 
\colorbox{red!42.0766348362142}{\strut P2} 
\colorbox{red!82.29179257826665}{\strut P3} 
\colorbox{red!66.37253716910432}{\strut P4} 
\colorbox{red!59.45847939078932}{\strut P5} 
\colorbox{red!40.11845763326484}{\strut P6}
\colorbox{red!62.23860751843346}{\strut P7} 
\colorbox{red!18.711471050404938}{\strut P8} 
\colorbox{red!43.15242354647649}{\strut P9} 
\colorbox{red!67.1703130666022}{\strut P10}
\colorbox{red!21.249848906080027}{\strut P11} 
\colorbox{red!46.077601837302073}{\strut P12} 
\colorbox{red!99.1296990209114}{\strut P13} 
\colorbox{red!21.34654901486765}{\strut P14} 
\colorbox{red!0.0}{\strut P15} 
\\
\colorbox{red!33.749406}{\strut T16}
& 
\colorbox{red!38.39199614271938}{\strut P0}
\colorbox{red!46.745419479267114}{\strut P1} 
\colorbox{red!56.81051108968177}{\strut P2} 
\colorbox{red!91.48987463837994}{\strut P3} 
\colorbox{red!27.08534233365477}{\strut P4} 
\colorbox{red!42.36981677917068}{\strut P5} 
\colorbox{red!85.14946962391514}{\strut P6}
\colorbox{red!100.0}{\strut P7} 
\colorbox{red!32.449373191899705}{\strut P8} 
\colorbox{red!78.09787849566057}{\strut P9} 
\colorbox{red!47.29990356798457}{\strut P10}
\colorbox{red!0.0}{\strut P11} 
\colorbox{red!29.315332690453233}{\strut P12} 
\colorbox{red!45.4556412729026}{\strut P13} 
\colorbox{red!48.686113789778207}{\strut P14} 
\colorbox{red!28.1702025072324}{\strut P15} 
\\
\colorbox{red!84.201141}{\strut T17}  
& 
\colorbox{red!87.63218853862422}{\strut P0}
\colorbox{red!100.0}{\strut P1} 
\colorbox{red!92.68808540638533}{\strut P2} 
\colorbox{red!59.71396918118642}{\strut P3} 
\colorbox{red!71.53791922650821}{\strut P4} 
\colorbox{red!23.627757075234168}{\strut P5} 
\colorbox{red!66.1899486353107}{\strut P6}
\colorbox{red!62.08077349179172}{\strut P7} 
\colorbox{red!35.70349481317353}{\strut P8} 
\colorbox{red!66.63309497431765}{\strut P9} 
\colorbox{red!75.3852351697049}{\strut P10}
\colorbox{red!63.67207170913486}{\strut P11} 
\colorbox{red!47.174942088830696}{\strut P12} 
\colorbox{red!0.0}{\strut P13} 
\colorbox{red!70.3293383019438}{\strut P14} 
\colorbox{red!64.88065263369927}{\strut P15} 
\\

\colorbox{red!23.240609}{\strut T18} 
& 
\colorbox{red!70.14836085187843}{\strut P0}
\colorbox{red!77.96123474515434}{\strut P1} 
\colorbox{red!0.0}{\strut P2} 
\colorbox{red!45.166307729121796}{\strut P3} 
\colorbox{red!100.0}{\strut P4} 
\colorbox{red!54.30725053840632}{\strut P5} 
\colorbox{red!87.56879636276621}{\strut P6}
\colorbox{red!36.12108159846853}{\strut P7} 
\colorbox{red!28.715003589375443}{\strut P8} 
\colorbox{red!59.8588178990189}{\strut P9} 
\colorbox{red!93.35965541995692}{\strut P10}
\colorbox{red!12.682459918640815}{\strut P11} 
\colorbox{red!19.215123235223736}{\strut P12} 
\colorbox{red!36.515912897822445}{\strut P13} 
\colorbox{red!19.526202440775303}{\strut P14} 
\colorbox{red!73.21129456807849}{\strut P15} 
\\
\colorbox{red!28.019496}{\strut T19} 
& 
\colorbox{red!19.85807614327551}{\strut P0}
\colorbox{red!100.0}{\strut P1} 
\colorbox{red!51.72336111736878}{\strut P2} 
\colorbox{red!66.2987159270106}{\strut P3} 
\colorbox{red!74.18337463392656}{\strut P4} 
\colorbox{red!65.69047082676278}{\strut P5} 
\colorbox{red!05.203874746564535}{\strut P6}
\colorbox{red!33.40842532101825}{\strut P7} 
\colorbox{red!44.886235638657357}{\strut P8} 
\colorbox{red!49.831043027708943}{\strut P9} 
\colorbox{red!26.875422392430726}{\strut P10}
\colorbox{red!36.62987159270106}{\strut P11} 
\colorbox{red!29.03807163775625}{\strut P12} 
\colorbox{red!28.53120072088308}{\strut P13} 
\colorbox{red!78.33971615228655}{\strut P14} 
\colorbox{red!0.0}{\strut P15}  \\
\bottomrule
\end{tabular}}
\caption{Importance scores visualization on WSC task. Among them, the shade of the red color indicates the level of the importance score, and the darker the color, the higher the importance score of the corresponding structure (token or piece). T$_i$ in first column denotes the $i$-th prompt token. P$_i$ in each row denotes the $i$-th prompt token piece.}
 \label{visual_importance_scores_piece_original}    
\end{figure}

\begin{figure}[!h]
\centering
\resizebox{0.44\textwidth}{!}{
\begin{tabular}{cc}
\toprule  
\textbf{Tokens} & \textbf{Soft Prompt Token Pieces}     \\  
\hline
\colorbox{red!63.552068}{\strut T0} 
& 
\colorbox{red!89.36}{\strut P0}
\colorbox{gray!50.0}{\strut P1} 
\colorbox{red!100.0}{\strut P2} 
\colorbox{gray!50.0}{\strut P3} 
\colorbox{gray!50.0}{\strut P4} 
\colorbox{gray!50.0}{\strut P5} 
\colorbox{gray!50.0}{\strut P6}
\colorbox{red!73.51}{\strut P7} 
\colorbox{gray!50.0}{\strut P8} 
\colorbox{gray!50.0}{\strut P9} 
\colorbox{gray!50.0}{\strut P10}
\colorbox{gray!50.0}{\strut P11} 
\colorbox{gray!50.0}{\strut P12} 
\colorbox{red!70.84}{\strut P13} 
\colorbox{gray!50.0}{\strut P14} 
\colorbox{gray!50.0}{\strut P15} 

\\ 
\colorbox{gray!50.0}{\strut T1} 
& 
\colorbox{gray!50.0}{\strut P0}
\colorbox{gray!50.0}{\strut P1} 
\colorbox{gray!50.0}{\strut P2} 
\colorbox{gray!50.0}{\strut P3} 
\colorbox{gray!50.0}{\strut P4} 
\colorbox{gray!50.0}{\strut P5} 
\colorbox{gray!50.0}{\strut P6}
\colorbox{gray!50.0}{\strut P7} 
\colorbox{gray!50.0}{\strut P8} 
\colorbox{gray!50.0}{\strut P9} 
\colorbox{gray!50.0}{\strut P10}
\colorbox{gray!50.0}{\strut P11} 
\colorbox{gray!50.0}{\strut P12} 
\colorbox{gray!50.0}{\strut P13} 
\colorbox{gray!50.0}{\strut P14} 
\colorbox{gray!50.0}{\strut P15} 
\\
\colorbox{red!61.578697}{\strut T2} 
& 
\colorbox{gray!50.0}{\strut P0}
\colorbox{gray!50.0}{\strut P1} 
\colorbox{gray!50.0}{\strut P2} 
\colorbox{gray!50.0}{\strut P3} 
\colorbox{gray!50.0}{\strut P4} 
\colorbox{gray!50.0}{\strut P5} 
\colorbox{red!87.46}{\strut P6}
\colorbox{gray!50.0}{\strut P7} 
\colorbox{gray!50.0}{\strut P8} 
\colorbox{gray!50.0}{\strut P9} 
\colorbox{red!71.28085821377794}{\strut P10}
\colorbox{red!99.99999999999998}{\strut P11} 
\colorbox{red!99.84490112446684}{\strut P12} 
\colorbox{gray!50.0}{\strut P13} 
\colorbox{gray!50.0}{\strut P14} 
\colorbox{gray!50.0}{\strut P15} 
\\
\colorbox{red!32.810271}{\strut T3}
& 
\colorbox{red!85.41362174802689}{\strut P0}
\colorbox{gray!50.0}{\strut P1} 
\colorbox{gray!50.0}{\strut P2} 
\colorbox{red!77.91094221962389}{\strut P3} 
\colorbox{red!99.34716944363247}{\strut P4} 
\colorbox{gray!50.0}{\strut P5} 
\colorbox{gray!50.0}{\strut P6}
\colorbox{red!100.0}{\strut P7} 
\colorbox{gray!50.0}{\strut P8} 
\colorbox{gray!50.0}{\strut P9} 
\colorbox{gray!50.0}{\strut P10}
\colorbox{gray!50.0}{\strut P11} 
\colorbox{gray!50.0}{\strut P12} 
\colorbox{gray!50.0}{\strut P13} 
\colorbox{gray!50.0}{\strut P14} 
\colorbox{gray!50.0}{\strut P15} 
\\
\colorbox{red!32.477413}{\strut T4}
& 
\colorbox{gray!50.0}{\strut P0}
\colorbox{red!62.31875038894767}{\strut P1} 
\colorbox{gray!50.0}{\strut P2} 
\colorbox{red!100.0}{\strut P3} 
\colorbox{gray!50.0}{\strut P4} 
\colorbox{gray!50.0}{\strut P5} 
\colorbox{gray!50.0}{\strut P6}
\colorbox{gray!50.0}{\strut P7} 
\colorbox{gray!50.0}{\strut P8} 
\colorbox{gray!50.0}{\strut P9} 
\colorbox{red!47.99303005787542}{\strut P10}
\colorbox{gray!50.0}{\strut P11} 
\colorbox{gray!50.0}{\strut P12} 
\colorbox{red!43.18216975009557}{\strut P13} 
\colorbox{gray!50.0}{\strut P14} 
\colorbox{gray!50.0}{\strut P15} 
\\
\colorbox{red!50.237756}{\strut T5}
& 
\colorbox{gray!50.0}{\strut P0}
\colorbox{red!91.04856736435685}{\strut P1} 
\colorbox{red!59.62202804308068}{\strut P2} 
\colorbox{red!60.07925218451534}{\strut P3} 
\colorbox{gray!50.0}{\strut P4} 
\colorbox{gray!50.0}{\strut P5} 
\colorbox{gray!50.0}{\strut P6}
\colorbox{red!100.0}{\strut P7} 
\colorbox{gray!50.0}{\strut P8} 
\colorbox{gray!50.0}{\strut P9} 
\colorbox{gray!50.0}{\strut P10}
\colorbox{gray!50.0}{\strut P11} 
\colorbox{gray!50.0}{\strut P12} 
\colorbox{gray!50.0}{\strut P13} 
\colorbox{gray!50.0}{\strut P14} 
\colorbox{gray!50.0}{\strut P15} 
\\
\colorbox{red!38.456966}{\strut T6}
& 
\colorbox{gray!50.0}{\strut P0}
\colorbox{gray!50.0}{\strut P1} 
\colorbox{red!89.7327532628962}{\strut P2} 
\colorbox{gray!50.0}{\strut P3} 
\colorbox{gray!50.0}{\strut P4} 
\colorbox{gray!50.0}{\strut P5} 
\colorbox{gray!50.0}{\strut P6}
\colorbox{gray!50.0}{\strut P7} 
\colorbox{gray!50.0}{\strut P8} 
\colorbox{gray!50.0}{\strut P9} 
\colorbox{gray!50.0}{\strut P10}
\colorbox{gray!50.0}{\strut P11}  
\colorbox{red!100.0}{\strut P12} 
\colorbox{red!93.53635798632691}{\strut P13} 
\colorbox{red!77.83716594157861}{\strut P14} 
\colorbox{gray!50.0}{\strut P15} 
\\
\colorbox{red!77.104137}{\strut T7}
& 
\colorbox{red!95.56994160954556}{\strut P0}
\colorbox{gray!50.0}{\strut P1} 
\colorbox{gray!50.0}{\strut P2} 
\colorbox{gray!50.0}{\strut P3} 
\colorbox{gray!50.0}{\strut P4} 
\colorbox{gray!50.0}{\strut P5} 
\colorbox{red!96.4077176948464}{\strut P6}
\colorbox{gray!50.0}{\strut P7} 
\colorbox{red!86.17669459253616}{\strut P8} 
\colorbox{gray!50.0}{\strut P9} 
\colorbox{gray!50.0}{\strut P10}
\colorbox{gray!50.0}{\strut P11} 
\colorbox{gray!50.0}{\strut P12} 
\colorbox{gray!50.0}{\strut P13} 
\colorbox{gray!50.0}{\strut P14}
\colorbox{red!99.99999999999999}{\strut P15} 
\\
\colorbox{gray!50.0}{\strut T8}
& 
\colorbox{gray!50.0}{\strut P0}
\colorbox{gray!50.0}{\strut P1} 
\colorbox{gray!50.0}{\strut P2} 
\colorbox{gray!50.0}{\strut P3} 
\colorbox{gray!50.0}{\strut P4} 
\colorbox{gray!50.0}{\strut P5} 
\colorbox{gray!50.0}{\strut P6}
\colorbox{gray!50.0}{\strut P7} 
\colorbox{gray!50.0}{\strut P8} 
\colorbox{gray!50.0}{\strut P9} 
\colorbox{gray!50.0}{\strut P10}
\colorbox{gray!50.0}{\strut P11} 
\colorbox{gray!50.0}{\strut P12} 
\colorbox{gray!50.0}{\strut P13} 
\colorbox{gray!50.0}{\strut P14} 
\colorbox{gray!50.0}{\strut P15} 
\\
\colorbox{red!59.938184}{\strut T9}
& 
\colorbox{red!78.78837228238743}{\strut P0}
\colorbox{gray!50.0}{\strut P1} 
\colorbox{gray!50.0}{\strut P2} 
\colorbox{gray!50.0}{\strut P3} 
\colorbox{gray!50.0}{\strut P4} 
\colorbox{gray!50.0}{\strut P5} 
\colorbox{gray!50.0}{\strut P6}
\colorbox{gray!50.0}{\strut P7} 
\colorbox{red!71.48440680726326}{\strut P8} 
\colorbox{gray!50.0}{\strut P9} 
\colorbox{red!100.0}{\strut P10}
\colorbox{gray!50.0}{\strut P11} 
\colorbox{gray!50.0}{\strut P12} 
\colorbox{gray!50.0}{\strut P13} 
\colorbox{red!81.31259669408029}{\strut P14} 
\colorbox{gray!50.0}{\strut P15} 
\\
\colorbox{gray!50.0}{\strut T10}
& 
\colorbox{gray!50.0}{\strut P0}
\colorbox{gray!50.0}{\strut P1} 
\colorbox{gray!50.0}{\strut P2} 
\colorbox{gray!50.0}{\strut P3} 
\colorbox{gray!50.0}{\strut P4} 
\colorbox{gray!50.0}{\strut P5} 
\colorbox{gray!50.0}{\strut P6}
\colorbox{gray!50.0}{\strut P7} 
\colorbox{gray!50.0}{\strut P8} 
\colorbox{gray!50.0}{\strut P9} 
\colorbox{gray!50.0}{\strut P10}
\colorbox{gray!50.0}{\strut P11} 
\colorbox{gray!50.0}{\strut P12} 
\colorbox{gray!50.0}{\strut P13} 
\colorbox{gray!50.0}{\strut P14} 
\colorbox{gray!50.0}{\strut P15} 
\\
\colorbox{gray!50.0}{\strut T11} 
& 
\colorbox{gray!50.0}{\strut P0}
\colorbox{gray!50.0}{\strut P1} 
\colorbox{gray!50.0}{\strut P2} 
\colorbox{gray!50.0}{\strut P3} 
\colorbox{gray!50.0}{\strut P4} 
\colorbox{gray!50.0}{\strut P5} 
\colorbox{gray!50.0}{\strut P6}
\colorbox{gray!50.0}{\strut P7} 
\colorbox{gray!50.0}{\strut P8} 
\colorbox{gray!50.0}{\strut P9} 
\colorbox{gray!50.0}{\strut P10}
\colorbox{gray!50.0}{\strut P11} 
\colorbox{gray!50.0}{\strut P12} 
\colorbox{gray!50.0}{\strut P13} 
\colorbox{gray!50.0}{\strut P14} 
\colorbox{gray!50.0}{\strut P15} 
\\
\colorbox{gray!50.0}{\strut T12}
& 
\colorbox{gray!50.0}{\strut P0}
\colorbox{gray!50.0}{\strut P1} 
\colorbox{gray!50.0}{\strut P2} 
\colorbox{gray!50.0}{\strut P3} 
\colorbox{gray!50.0}{\strut P4} 
\colorbox{gray!50.0}{\strut P5} 
\colorbox{gray!50.0}{\strut P6}
\colorbox{gray!50.0}{\strut P7} 
\colorbox{gray!50.0}{\strut P8} 
\colorbox{gray!50.0}{\strut P9} 
\colorbox{gray!50.0}{\strut P10}
\colorbox{gray!50.0}{\strut P11} 
\colorbox{gray!50.0}{\strut P12} 
\colorbox{gray!50.0}{\strut P13} 
\colorbox{gray!50.0}{\strut P14} 
\colorbox{gray!50.0}{\strut P15} 
\\
\colorbox{red!57.310984}{\strut T13} 
& 
\colorbox{gray!50.0}{\strut P0}
\colorbox{gray!50.0}{\strut P1} 
\colorbox{red!96.30625248904819}{\strut P2} 
\colorbox{gray!50.0}{\strut P3} 
\colorbox{gray!50.0}{\strut P4} 
\colorbox{gray!50.0}{\strut P5} 
\colorbox{gray!50.0}{\strut P6}
\colorbox{red!83.92074870569495}{\strut P7} 
\colorbox{red!100.0}{\strut P8} 
\colorbox{gray!50.0}{\strut P9} 
\colorbox{gray!50.0}{\strut P10}
\colorbox{gray!50.0}{\strut P11} 
\colorbox{red!98.40700915969733}{\strut P12} 
\colorbox{gray!50.0}{\strut P13} 
\colorbox{gray!50.0}{\strut P14} 
\colorbox{gray!50.0}{\strut P15} 
\\
\colorbox{red!100.0}{\strut T14}
& 
\colorbox{gray!50.0}{\strut P0}
\colorbox{gray!50.0}{\strut P1} 
\colorbox{red!91.91893278604413}{\strut P2} 
\colorbox{gray!50.0}{\strut P3} 
\colorbox{gray!50.0}{\strut P4} 
\colorbox{red!72.90918419702412}{\strut P5} 
\colorbox{gray!50.0}{\strut P6}
\colorbox{red!98.51205746536686}{\strut P7} 
\colorbox{gray!50.0}{\strut P8} 
\colorbox{gray!50.0}{\strut P9} 
\colorbox{gray!50.0}{\strut P10}
\colorbox{gray!50.0}{\strut P11} 
\colorbox{gray!50.0}{\strut P12}  
\colorbox{red!100.0}{\strut P13} 
\colorbox{gray!50.0}{\strut P14} 
\colorbox{gray!50.0}{\strut P15} 
\\
\colorbox{gray!50.0}{\strut T15} 
& 
\colorbox{gray!50.0}{\strut P0}
\colorbox{gray!50.0}{\strut P1} 
\colorbox{gray!50.0}{\strut P2} 
\colorbox{gray!50.0}{\strut P3} 
\colorbox{gray!50.0}{\strut P4} 
\colorbox{gray!50.0}{\strut P5} 
\colorbox{gray!50.0}{\strut P6}
\colorbox{gray!50.0}{\strut P7} 
\colorbox{gray!50.0}{\strut P8} 
\colorbox{gray!50.0}{\strut P9} 
\colorbox{gray!50.0}{\strut P10}
\colorbox{gray!50.0}{\strut P11} 
\colorbox{gray!50.0}{\strut P12} 
\colorbox{gray!50.0}{\strut P13} 
\colorbox{gray!50.0}{\strut P14} 
\colorbox{gray!50.0}{\strut P15} 
\\
\colorbox{red!33.749406}{\strut T16}
& 
\colorbox{gray!50.0}{\strut P0}
\colorbox{gray!50.0}{\strut P1} 
\colorbox{gray!50.0}{\strut P2} 
\colorbox{red!91.48987463837994}{\strut P3} 
\colorbox{gray!50.0}{\strut P4} 
\colorbox{gray!50.0}{\strut P5} 
\colorbox{red!85.14946962391514}{\strut P6}
\colorbox{red!100.0}{\strut P7} 
\colorbox{gray!50.0}{\strut P8} 
\colorbox{red!78.09787849566057}{\strut P9} 
\colorbox{gray!50.0}{\strut P10}
\colorbox{gray!50.0}{\strut P11} 
\colorbox{gray!50.0}{\strut P12} 
\colorbox{gray!50.0}{\strut P13} 
\colorbox{gray!50.0}{\strut P14} 
\colorbox{gray!50.0}{\strut P15}
\\
\colorbox{red!84.201141}{\strut T17}  
& 
\colorbox{red!87.63218853862422}{\strut P0}
\colorbox{red!100.0}{\strut P1} 
\colorbox{red!92.68808540638533}{\strut P2} 
\colorbox{gray!50.0}{\strut P3} 
\colorbox{gray!50.0}{\strut P4} 
\colorbox{gray!50.0}{\strut P5} 
\colorbox{gray!50.0}{\strut P6}
\colorbox{gray!50.0}{\strut P7} 
\colorbox{gray!50.0}{\strut P8} 
\colorbox{gray!50.0}{\strut P9} 
\colorbox{red!75.3852351697049}{\strut P10}
\colorbox{gray!50.0}{\strut P11} 
\colorbox{gray!50.0}{\strut P12} 
\colorbox{gray!50.0}{\strut P13} 
\colorbox{gray!50.0}{\strut P14} 
\colorbox{gray!50.0}{\strut P15} 
\\

\colorbox{gray!50.0}{\strut T18} 
& 
\colorbox{gray!50.0}{\strut P0}
\colorbox{gray!50.0}{\strut P1} 
\colorbox{gray!50.0}{\strut P2} 
\colorbox{gray!50.0}{\strut P3} 
\colorbox{gray!50.0}{\strut P4} 
\colorbox{gray!50.0}{\strut P5} 
\colorbox{gray!50.0}{\strut P6}
\colorbox{gray!50.0}{\strut P7} 
\colorbox{gray!50.0}{\strut P8} 
\colorbox{gray!50.0}{\strut P9} 
\colorbox{gray!50.0}{\strut P10}
\colorbox{gray!50.0}{\strut P11} 
\colorbox{gray!50.0}{\strut P12} 
\colorbox{gray!50.0}{\strut P13} 
\colorbox{gray!50.0}{\strut P14} 
\colorbox{gray!50.0}{\strut P15} 
\\
\colorbox{gray!50.0}{\strut T19} 
& 
\colorbox{gray!50.0}{\strut P0}
\colorbox{gray!50.0}{\strut P1} 
\colorbox{gray!50.0}{\strut P2} 
\colorbox{gray!50.0}{\strut P3} 
\colorbox{gray!50.0}{\strut P4} 
\colorbox{gray!50.0}{\strut P5} 
\colorbox{gray!50.0}{\strut P6}
\colorbox{gray!50.0}{\strut P7} 
\colorbox{gray!50.0}{\strut P8} 
\colorbox{gray!50.0}{\strut P9} 
\colorbox{gray!50.0}{\strut P10}
\colorbox{gray!50.0}{\strut P11} 
\colorbox{gray!50.0}{\strut P12} 
\colorbox{gray!50.0}{\strut P13} 
\colorbox{gray!50.0}{\strut P14} 
\colorbox{gray!50.0}{\strut P15}  \\
\bottomrule
\end{tabular}}
\caption{Importance scores visualization after \textsc{XPrompt} on WSC task. The gray elements indicate that the prompt tokens or pieces are pruned due to low importance scores, and the remaining parts are the positive tokens or positive pieces.}
 \label{visual_importance_scores_piece_masked}    
\end{figure}

\section{SuperGLUE Statistics, Metrics and Soft Prompt Templates}
\label{section:SuperGLUE_Statistics_and_Metrics}

\paragraph{SuperGLUE benchmark} is a collection of eight challenging language understanding tasks designed to be summarized into a single metric, including question answering (BoolQ \citep{clark2019boolq}, MultiRC \citep{khashabi2018looking}, ReCoRD \citep{zhang2018record}), textual entailment (RTE \citep{dagan2005pascal}, CB \citep{clark2019boolq}), coreference resolution (WSC \citep{levesque2012winograd}), word sense disambiguation (WiC \citep{pilehvar2019wic}), and causal reasoning (COPA \citep{roemmele2011choice}). Following previous works \citep{schick2021s, liu2021gpt}, we focus on $7$ of them, excepting ReCoRD task, since the ReCoRD is also QA tasks. The detailed statistics and metrics are provided in Table~\ref{SuperGLUE}, and the soft prompt templates and generation verbalizers are provided in Table~\ref{SoftTemplate}.


\begin{table*}[ht]
\centering
\resizebox{0.85\textwidth}{!}{
\begin{tabular}{ccccccl}
\toprule
Dataset  & Train & Dev  & Test & Task        & Metrics & \multicolumn{1}{c}{Text Sources} \\ \hline
BoolQ   & 9427  & 3270 & 3245 & QA      & Acc     & Google queries, Wikipedia  \\
CB      & 250   & 57   & 250  & NLI     & Acc  & Various                    \\
COPA    & 400   & 100  & 500  & QA      & Acc     & Blogs, Photography encyclopedia \\
MultiRC & 5100  & 953  & 1800 & QA      &F1$_{a}$  & Various          \\           
RTE     & 2500  & 278  & 300  & NLI     & Acc     & News, Wikipedia         \\
WiC     & 6000  & 638  & 1400 & WSD     & Acc     & WordNet, VerbNet, Wiktionary \\
WSC     & 554   & 104  & 146  & Coreference & Acc     & Fiction books               \\ 
\bottomrule
\end{tabular}}
 \caption{The data statistics and metrics of seven SuperGLUE tasks. WSD stands for word sense disambiguation, NLI is natural language inference, Coreference is coreference resolution, and QA is question answering. Acc is accuracy, and F1$_{a}$ is F1-score over all answer-options.}
 \label{SuperGLUE}
\end{table*}

\begin{table*}[]
\centering
\resizebox{0.98\textwidth}{!}{
\begin{tabular}{cccc}
\toprule
Dataset &Task &\multicolumn{1}{c}{Soft Template}                               & \multicolumn{1}{c}{Generation Verbalizers} \\ \hline
BoolQ   & QA & \begin{tabular}[c]{@{}l@{}}\{Soft Prompt Tokens\} hypothesis: \{"placeholder":"text\_b", "shortenable":False, \\ "post\_processing": lambda x:x+"."\} premise: \{"placeholder":"text\_a"\} \{"mask"\} \end{tabular}                                                  & "yes" / "no"                          \\ \hline     
CB      & NLI         & \begin{tabular}[c]{@{}l@{}}\{Soft Prompt Tokens\} hypothesis: \{"placeholder":"text\_b","post\_processing": \\ lambda x:x+"."\} premise: \{"placeholder":"text\_a"\} \{"mask"\}\end{tabular}                                                                        & "entailment" / "contradiction" / "neutral" \\ \hline     
COPA    & QA          & \begin{tabular}[c]{@{}l@{}}\{Soft Prompt Tokens\} choice1: \{"meta":"choice1"\} choice2: \{"meta":"choice2"\}\\  premise: \{"placeholder":"text\_a"\} question: \{"meta":"question"\} \{"mask"\}\end{tabular}                                                       & "choice1" / "choice2"                      \\ \hline     
MultiRC & QA          & \begin{tabular}[c]{@{}l@{}}\{Soft Prompt Tokens\} question: \{"placeholder":"text\_b", "shortenable":False\} answer: \{"meta":"answer", \\ "shortenable":False, "post\_processing": lambda x:x+"."\} paragraph: \{"placeholder":"text\_a"\} \{"mask"\}\end{tabular} & "yes" / "no"                               \\ \hline     
RTE     & NLI         & \{Soft Prompt Token\} sentence1: \{"placeholder":"text\_a"\} \\
& & sentence2: \{"placeholder":"text\_b"\} \{"mask"\}                                                                                                                                                     & "entailment" / "contradiction"             \\ \hline     
WiC     & WSD         & \begin{tabular}[c]{@{}l@{}}\{Soft Prompt Tokens\} sentence1: \{"placeholder":"text\_a"\} sentence2: \\ \{"placeholder":"text\_b"\} word: \{"meta":"word", "shortenable": False\} \{"mask"\}\end{tabular}                                                            & "yes" / "no"                               \\ \hline     
WSC     & Coreference & \begin{tabular}[c]{@{}l@{}}\{Soft Prompt Tokens\} \{"placeholder":"text\_a"\} "\{"meta":"span2\_text"\}"\\  refers to "\{"meta":"span1\_text"\}" or another word ? \{"mask"\}\end{tabular}                                                                          & "another word" / "span1\_text"             \\ \bottomrule
\end{tabular}}
\caption{The soft prompt templates and generation verbalizers for the seven SuperGLUE tasks used in our experiments.}
 \label{SoftTemplate}
\end{table*}

\end{document}